\documentclass[sigconf]{acmart}

\AtBeginDocument{%
  }

\copyrightyear{2023}
\acmYear{2023}
\setcopyright{acmlicensed}\acmConference[KDD '23]{Proceedings of the 29th ACM SIGKDD Conference on Knowledge Discovery and Data Mining}{August 6--10, 2023}{Long Beach, CA, USA}
\acmBooktitle{Proceedings of the 29th ACM SIGKDD Conference on Knowledge Discovery and Data Mining (KDD '23), August 6--10, 2023, Long Beach, CA, USA}
\acmPrice{15.00}
\acmDOI{10.1145/3580305.3599541}
\acmISBN{979-8-4007-0103-0/23/08}

\usepackage{amsmath,amsfonts,bm}

\def\eqref#1{equation~\ref{#1}}

\def\1{\bm{1}}

\def\ve{{\bm{e}}}

\def\vp{{\bm{p}}}

\def\vx{{\bm{x}}}

\def\mA{{\bm{A}}}

\def\mE{{\bm{E}}}

\def\mL{{\bm{L}}}

\def\mP{{\bm{P}}}

\def\mX{{\bm{X}}}

\def\mZ{{\bm{Z}}}

\DeclareMathAlphabet{\mathsfit}{\encodingdefault}{\sfdefault}{m}{sl}
\SetMathAlphabet{\mathsfit}{bold}{\encodingdefault}{\sfdefault}{bx}{n}

\def\gG{{\mathcal{G}}}

\def\sC{{\mathbb{C}}}

\def\sN{{\mathbb{N}}}

\def\sQ{{\mathbb{Q}}}
\def\sR{{\mathbb{R}}}
\def\sS{{\mathbb{S}}}

\newcommand{\R}{\mathbb{R}}

\usepackage{balance}
\usepackage{caption}
\usepackage{subcaption}
\usepackage{hyperref}
\usepackage{url}
\usepackage{graphicx}
\usepackage{enumitem}
\newtheorem{definition}{Definition}

\setcopyright{acmcopyright}

\usepackage{algorithm}
\usepackage{algorithmic}
\usepackage{booktabs}
\usepackage{wrapfig,lipsum}
\usepackage{wrapfig}
\usepackage{multirow}

\usepackage[utf8]{inputenc}
\begin{document}


\title{Virtual Node Tuning for Few-shot Node Classification}

\author{Zhen Tan}
\affiliation{%
  \institution{Arizona State University}
  \country{}
}
\email{ztan36@asu.edu}

\author{Ruocheng Guo}
\affiliation{%
  \institution{ByteDance Research}
  \country{}
  }
\email{ruocheng.guo@bytedance.com}

\author{Kaize Ding}
\affiliation{%
  \institution{Arizona State University}
  \country{}
}
\email{kaize.ding@asu.edu}

\author{Huan Liu}
\affiliation{%
  \institution{Arizona State University}
  \country{}
}
\email{huanliu@asu.edu}

\renewcommand{\shortauthors}{Zhen Tan, Ruocheng Guo, Kaize Ding, \& Huan Liu}

\begin{abstract}
  Few-shot Node Classification (FSNC) is a challenge in graph representation learning where only a few labeled nodes per class are available for training. To tackle this issue, meta-learning has been proposed to transfer structural knowledge from base classes with abundant labels to target novel classes. However, existing solutions become ineffective or inapplicable when base classes have no or limited labeled nodes. To address this challenge, we propose an innovative method dubbed Virtual Node Tuning (VNT). Our approach utilizes a pretrained graph transformer as the encoder and injects virtual nodes as soft prompts in the embedding space, which can be optimized with few-shot labels in novel classes to modulate node embeddings for each specific FSNC task. A unique feature of VNT is that, by incorporating a Graph-based Pseudo Prompt Evolution (GPPE) module, VNT-GPPE can handle scenarios with sparse labels in base classes. Experimental results on four datasets demonstrate the superiority of the proposed approach in addressing FSNC with unlabeled or sparsely labeled base classes, outperforming existing state-of-the-art methods and even fully supervised baselines. 
\end{abstract}




\maketitle
\setcounter{page}{1}
\section{Introduction}
With the advent of deep learning, Graph Neural Networks (GNNs) have been proposed for effective graph representation learning with sufficient labeled instances~\citep{Kipf:2017tc,Hamilton:2017tp,Velickovic:2018we,xu2018powerful}. However, there is a growing interest in learning GNNs with limited labels, which is a prevalent issue in large graphs where manual data collection and labeling is costly~\cite{ding2022data}. This has led to a proliferation of studies in the field of Few-shot Node Classification (FSNC)~\citep{zhang2018few,ding2020graph,wang21AMM,huang2020graph,lan2020node,wang2022task}, which aims to learn fast-adaptable GNNs for unseen tasks with extremely scarce ground-truth labels. Conventionally, FSNC tasks are denoted as $N$-way $K$-shot $R$-query node classification tasks, where $N$ is the number of classes, $K$ is the number of labeled nodes per class, and $R$ is the number of unlabeled nodes per class. The labeled nodes are referred to as the "support set" and the unlabeled nodes are referred to as the "query set" for evaluation.

The current modus operandi, i.e., \textit{meta-learning}, has become a predominate and successful paradigm to tackle the label scarcity issue on graphs~\citep{zhang2018few,ding2020graph,huang2020graph,wang2022task,tan2022graph}. Besides the target node classes (termed as \textit{novel classes}) with few labeled nodes, meta-learning-based methods assume the existence of a set of \textit{base classes}, which is disjoint with the novel classes set and has substantial labeled nodes in each class to sample a number of meta-tasks, or episodes, to train the GNN model while emulating the $N$-way $K$-shot $R$-query task structure. This emulation-based training has been proved helpful for fast adaptation to target FSNC tasks \citep{lan2020node,wang21AMM}. Despite astonishing breakthroughs having been made, \citet{tan2022simple} firstly points out that those meta-learning-based methods suffer from the piecemeal graph knowledge issue, which implies that only a small portion of nodes are involved in each episode, thus hindering the generalizability of the learned GNN models regarding unseen novel classes. Additionally, the assumption of the existence of substantially labeled base classes may not be feasible for real-world graphs~\cite{tan2022transductive}. In summary, while meta-learning is a successful method for FSNC tasks, it has limitations in terms of effectiveness and applicability.

Considering these limitations in the existing efforts, in this work, we first generalize the traditional definition of FSNC tasks to cover more real-world scenarios where there could be limited or even no labels even in base classes. We first start with the most challenging setting where no available labeled nodes exist in base classes. To facilitate sufficient training, we choose \textit{Graph Transformers} (GTs)~\citep{zhang2020graph,chen2022structure} as the encoder to learn representative node embeddings. Recently, large transformer-based~\citep{vaswani2017attention} models have thrived in various domains, such as languages~\citep{devlin2018bert}, images~\citep{dosovitskiy2020image}, as well as graphs~\citep{hu2020heterogeneous}. The number of parameters of GTs can be much larger than traditional GNNs by orders of magnitude, which has shown unique advantages in modeling graph data and acquiring structural knowledge~\citep{zhang2020graph,chen2022structure}. Furthermore, pretrained in an unsupervised manner, GTs can learn from a large number of unlabeled nodes by enforcing the model to learn from pre-defined pretext tasks (e.g. masked link restoration, masked node recovery, etc.)~\citep{zhang2020graph,hu2020heterogeneous}. In other words, no node label information from base classes is needed for obtaining pretrained GTs enriched with topological and semantic knowledge. However, our experiments show that directly transferring node embedding from GTs and fine-tuning the GT encoder on the support set will lead to unsatisfactory performance. This is because directly transferring node embeddings neglects the inherent gap between the training objective of the pretexts and that of the downstream FSNC tasks. Also, naively fine-tuning with the few labeled nodes will lead to severe overfitting. Both these two factors can render the transferred node embeddings sub-optimal for target FSNC tasks. Accordingly, to elicit the learned substantial prior graph knowledge from GTs with only a few labels from each target task, we propose a method, \textit{Virtual Node Tuning} (VNT), that can efficiently modulate the GTs to customize the pretrained node embeddings for different FSNC tasks. 

Recent advancements in natural language processing (NLP) have led to the emergence of a new technique called "prompting" for adapting large-scale transformer-based language models to new few-shot or zero-shot tasks~\citep{liu2021pre}. It refers to prepending language instructions to the input text to guide those language models to better \textit{understand} the new task and give more tailored \textit{replies}. However, such a technique cannot be straightforwardly applied to GTs due to the significant disparity between graphs and texts. Given the symbolic nature of graph data, it is infeasible and counter-intuitive to manually design semantic prompts like human languages for each target FSNC task. Inspired by more recent works~\citep{lester2021power,jia2022visual}, instead of manually creating prompts in the raw graph data space (e.g., nodes and edges), we propose to inject a set of continuous vectors as task-specific virtual nodes in the node embedding space to function as \textit{soft prompts} to elicit the substantial knowledge contained in the learned GTs. During the fine-tuning phase, these prompts can be optimized via the few labeled nodes from the support set in each FSNC task. This simple tuning with virtual node prompts can modulate the learned node embeddings according to the context of the FSNC task. Moreover, for scenarios where sparsely labeled nodes exist in base classes, we propose to reformulate the problem by assuming the presence of a few \textit{source} FSNC tasks within the base class label space. Meanwhile, we find that initializing the prompt of an FSNC task as the prompt of a previously learned FSNC task can potentially yield positive transfer. Based on this observation, we design a novel \textit{Graph-based Pseudo Prompt Evolution} (GPPE) module, which performs a prompt-level meta-learning to selectively transfer knowledge learned from source FSNC tasks to target FSNC tasks. This module has demonstrated promising improvement for VNT and scales well to conditions where node labels are highly scarce, i.e., very few source tasks exist.

Notably, the proposed framework is fully automatic and requires no human involvement. By only retraining a small prompt tensor and a simple classifier, and recycling a single GT for all downstream FSNC tasks, our method significantly reduces storage and computation costs per task. Through extensive experimentation, we have demonstrated the effectiveness of the VNT-GPPE method
in terms of both accuracy and efficiency. We hope this work can provide a new promising path forward for \textit{few-shot Node Classification} (FSNC) tasks. In summary, our main contributions include:
\vspace{-0.05in}
\setlength\itemsep{0.2em}
\begin{itemize}[leftmargin=*]
    \item \textbf{Problem Generalization:} We relax the assumption in conventional FSNC tasks to cover scenarios where there could be none or sparsely labeled nodes in base classes.
    \item \textbf{Framework Proposed:} We propose a simple yet effective framework, \textit{Virtual Node Tuning} (VNT), that does not rely on any label from base classes. We inject virtual nodes in the embedding space which function as soft prompts to customize the pretrained node embeddings for each FSNC task. To extend the framework for scenarios where sparely labeled nodes in base classes are available, we further design a \textit{Graph-based Pseudo Prompt Evolution} (GPPE) module that transfers prompts learned from base classes to target downstream FSNC tasks.
    \item \textbf{Comprehensive Experiment:} We conduct extensive experiments on four widely-used real-world datasets to show the effectiveness and applicability of the proposed framework. We find that VNT achieves competitive performance even no labeled nodes from base classes are utilized. Given sparsely labeled base classes, VNT-GPPE outperforms all the baselines even if they are given fully labeled base classes. Further analysis also indicates that VNT considerably benefits from prompt ensemble.
\end{itemize}




\section{Problem Formulation}
\label{sec:problem}
\label{problem}

In this work, we focus on few-shot node classification (FSNC) tasks on a single graph. Formally, given an attributed network $\gG = (\mathcal{V}, \mathcal{E}, \mX) = (\mA, \mX)$, where $\mA = \{0, 1\}^{V\times V}$ is the adjacency matrix representing the network structure, $\mX = [\vx_1;\vx_2; ...;\vx_V]$ represents all the node features, $\mathcal{V}$ denotes the set of vertices $\{v_1, v_2, ..., v_V\}$, and $\mathcal{E}$ denotes the set of edges $\{e_1, e_2, ..., e_E\}$. Specifically, $\mA_{j,k} = 1$ indicates that there is an edge between node $v_j$ and node $v_k$; otherwise, $\mA_{j,k} = 0$. The few-shot node classification problem assumes the existence of a series of node classification tasks, $\mathcal{T} = \{\mathcal{T}_i\}^{I}_{i=1}$, where $\mathcal{T}_i$ denotes the given dataset of a task, $I$ denotes the number of such tasks. Traditional FSNC tasks assume that those tasks are formed from target \textit{novel classes} (i.e. $\sC_{novel}$), where only a few labeled nodes are available per class, and there exists a disjoint set of \textit{base classes} (i.e. $\sC_{base}$, $\sC_{base} \cap \sC_{novel} = \varnothing$) on the graph where substantial labeled nodes are accessible during training. Next, we first present the definition of an $N$-way $K$-shot $R$-query node classification task as follows:
\begin{definition}
\textbf{N-way K-shot R-query Node Classification:} Given an attributed graph $\gG = (\mA, \mX)$ with a specified node label space $\sC$, $|\sC|=N$. If for each class $c\in \sC$, there are $K$ labeled nodes (i.e. support set $\sS$) as references and another $R$ nodes (i.e. query set $\sQ$) for prediction, then we term this task as an N-way K-shot R-query Node Classification task.
\end{definition}
Then, the traditional few-shot node classification problem can be defined as follows:
\begin{definition}
\textbf{Traditional Few-shot Node Classification:} Given an attributed graph $\gG = (\mA, \mX)$ with a disjoint node label space $\sC = \{\sC_{base}, \sC_{novel}\}$. Substantial labeled nodes from $\sC_{base}$ are available for sampling an arbitrary number of $N$-way $K$-shot $R$-query Node Classification tasks for training. The goal is to perform $N$-way $K$-shot $R$-query Node Classification for tasks sampled from $\sC_{novel}$.
\end{definition}

However, the assumption of the existence of substantial labeled nodes in the base classes could be untenable for real-world graphs. For example, all the classes on a given graph may only have a few labeled nodes. Considering this limitation, in this paper, we generalize the definition of FSNC and reformulate it according to the label sparsity within base classes. It is formulated as follows: 

\begin{definition}\label{def:gfsnc}
\textbf{General Few-shot Node Classification:} Given an attributed graph $\gG = (\mA, \mX)$ with a disjoint node label space $\sC = \{\sC_{base}, \sC_{novel}\}$. For $\sC_{base}$, there are $M$ $N$-way $K$-shot $R$-query Node Classification tasks for training. The goal is to perform $N$-way $K$-shot $R$-query Node Classification for tasks sampled from $\sC_{novel}$.
\end{definition}

Note that the key difference of the general FSNC compared to the traditional counterparts lies in the introduced parameter $M$. Since $M,N,K,R \ll~|V|$, the labels in base classes can be very sparse and the value of $M$ determines the label sparsity in the base classes. For example, if $M=0$, then the training phase actually provides no label from base classes, so the training procedure should be fully unsupervised. If $M\neq 0$, that means we have a collection of tasks with labeled nodes in base classes. In practice, we achieve this by random sampling $M$ $N$-way $K$-shot $R$-query node classification tasks from base classes. We term those $M$ tasks as \textit{source tasks} and the tasks during final evaluation as \textit{target tasks}. Particularly, if $M$ is a very large number, then the general FSNC will scale to the traditional FSNC, which most of the existing works are trying to address. Conversely, if $M$ is a relatively small number (e.g. $48$), this signifies the labels provided in the base classes are very sparse, which is the usual scenario for real-world applications. 


Our paper is the first to propose this more general problem formulation for FSNC tasks. 
To tackle this problem, in this work, we propose a novel framework named \textit{Virtual Node Tuning} that achieves promising performance when no source task exists (i.e., $M=0$). To cover more real-world scenarios (i.e., $M\neq0$), we further design a \textit{Prompt Transferring} mechanism via \textit{Graph-based Pseudo Prompt Evolution} that performs a prompt-level meta-learning to effectively transfer generalizable knowledge from source tasks to target downstream FSNC tasks. 

\section{Methodology}
\subsection{Preliminary: Graph Transformers}
\textit{Graph Transformers} (GTs)~\citep{zhang2020graph,rong2020self,chen2022structure} are Graph Neural Networks (GNNs) based on transformer~\citep{vaswani2017attention} without relying on convolution or aggregation operations. Following BERT~\citep{devlin2018bert} for large-scale natural language modeling, a $D$-layer GT is used to project the node attribute $\vx_j$ of each node $v_j$ ($\forall j \in \sN, 1\leq j \leq V$) into the corresponding embedding $\ve_j$. GTs usually have much more parameters than traditional GNNs and are often trained in a self-supervised manner, without the need for substantial gold-labeled nodes. For the sake of generality, we choose two simplest and most universally-used pretext
tasks, \textit{node attribute reconstruction} and \textit{structure recovery}, to pretrain the GT encoder~\citep{zhang2020graph,chen2022structure}. An exhaustive discussion of methods for pretraining GTs is out of the scope of this paper, please see more details for GT pretraining in Appendix \ref{app:implement}.  
Then, with a pretrained GT, each node $v_j$ is projected into a $F$-dimensional embedding space. With both node attribute and topology (or position) structure considered, the embedding matrix of all nodes in the graph $\gG$ is:
\begin{equation}
    \mE^0 = [\ve_1^0;...;\ve_j^0; ...; \ve_V^0] = Embed(\gG) =  Embed (\mX, \mA) \in \R^{V\times F},
\end{equation}
where $n$ is the number of nodes in the given graph $\gG$ and $V$ is the embedding size.
Then, the node embeddings $\mE^{d-1}$ computed by the $d-1$-th layer are fed into the following transformer layer $L^d$ ($\forall d \in \sN, 1\leq d \leq D$) to get more high-level node representations, which can be formulated as:
\begin{equation}
    \mE^d = [\ve_1^d; ...; \ve_j^d; ...; \ve_V^d] = L^d (\mE^{d-1}) \in \R^{V\times F}. 
\end{equation}

Conventionally, to adapt the pretrained GT to different downstream tasks, further fine-tuning of the GT on the corresponding datasets~\citep{zhang2020graph,rong2020self,chen2022structure} is performed. However, according to our experiments in Section \ref{sec:exp}, this vanilla approach suffers from the following limitations when applied to FSNC tasks: \textbf{(1)} The number of labeled nodes for each FSNC task is very limited (usually less than 5 labels), making the fine-tuned GT highly overfit on them and hard to generalize to query set. \textbf{(2)} This method neglects the inherent gap between the training objective of the pretext tasks and that of the downstream FSNC tasks, rendering the transferred node embeddings sub-optimal for the target FSNC tasks. \textbf{(3)} For every new task, all the parameters of GT models need to be updated, making the model hard to converge and greatly raising the cost to apply GTs to real-world applications. \textbf{(4)} GTs are generally pretrained in an unsupervised manner. How to utilize the labels (which can be sparse) in base classes to extract generalizable graph patterns for GTs remains unresolved. This work is the first to propose a simple yet effective and efficient \textit{prompting} method for GTs to tackle the four aforementioned limitations.

\begin{figure*}[htbp]
  \centering\scalebox{1.}{
  \includegraphics[width=\linewidth]{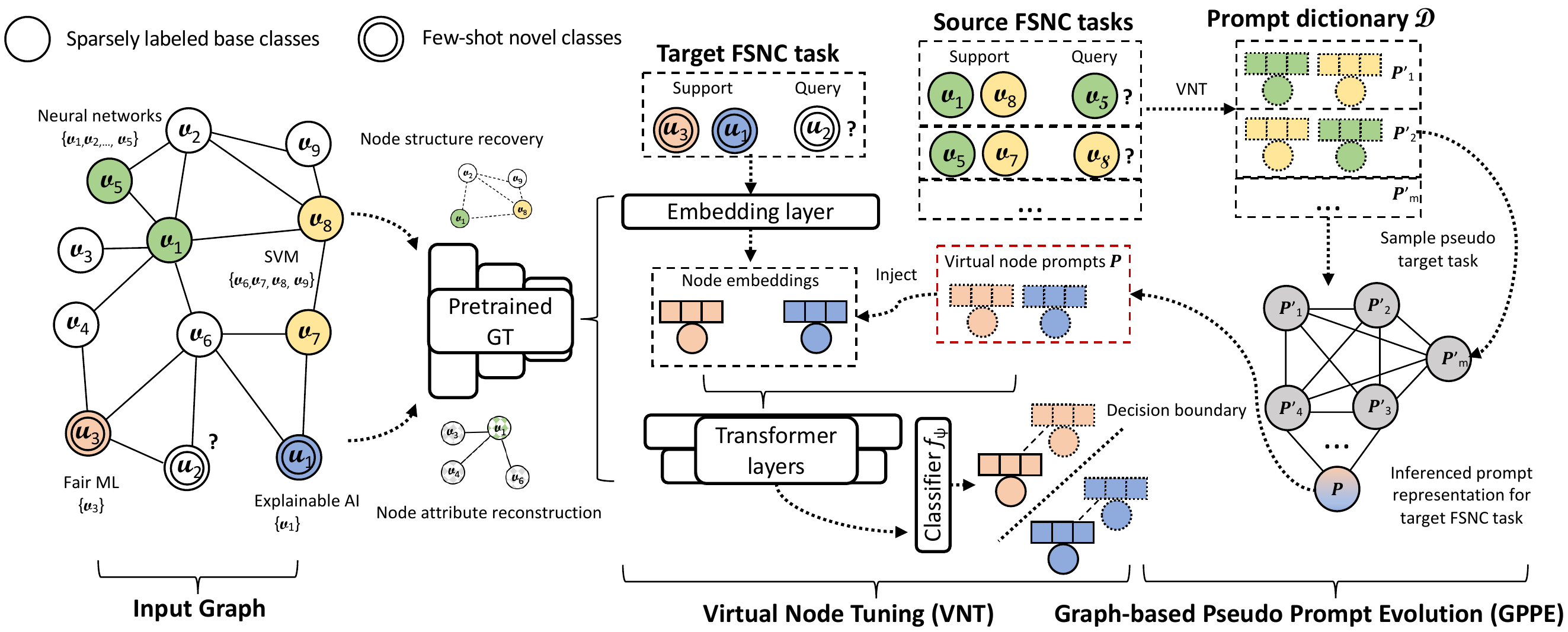}}
  \caption{The illustration of the proposed framework, VNT-GPPE. Colors indicate different classes (e.g., \textit{Neural Networks, SVM, Fair ML, Explainable AI}). Especially, white nodes mean labels of those nodes are unavailable. Different types of nodes indicate if nodes are from base classes or novel classes. Note that during VNT and GPPE, the parameters of the GT, including the embedding layer and transformer layers, are fixed.}
  \label{fig:framework}
\end{figure*}

\subsection{Virtual Node Tuning}\label{sec:NPT}
Since the GT encoder is pretrained on the given graph in a self-supervised manner, it does not require or utilize any label information from base classes. This characteristic fits the FSNC tasks where no source task exists, i.e., $M=0$. Then, in this section, we introduce the proposed \textit{Virtual Node Tuning} (VNT) method which effectively utilizes the limited few labeled nodes in the \textit{support set $\sS$} from the target label space $\sC$ to customize the node representations from the pretrained GT for each specific FSNC task. 

We introduce an extra set of $p$ randomly initialized continuous trainable vectors with the same embedding size $F$, \textit{i.e. prompt}, denoted as $\mP = [\vp_1;...;\vp_{p};...;\vp_P], (\vp_{p} \in \R^F)$. Our virtual node tuning strategy is simple to implement. We fix the pretrained weights of the GT encoder during fine-tuning while keeping the prompt parameter $\mP$ trainable, and we concatenate this prompt with the pretrained node embedding right after the embedding layer and feed it to the first transformer layer of the GT. The injected prompts can be viewed as task-specific \textit{virtual nodes} that help modulate the pretrained node representations and elicit the learned substantial knowledge from the pretrained GT for different target FSNC tasks. In such a manner, our approach allows the frozen large transformer layers to update the intermediate-layer node representations for different tasks, as contextualized by those virtual nodes (more detailed discussion about the effect from those virtual nodes on target FSNC tasks is presented in Section~\ref{exp:ablation} and \ref{exp:link}):
\begin{equation}
\label{eq:l1}
    [\mE^1 || \mZ^1] = L^1 ([\mE^{0} || \mP]) \in \R^{(V+P)\times F},
\end{equation}
where $||$ denotes the concatenation operator. Then we feed the learned node representations and the prompt to the following transformer layers $L^d$ ($\forall d \in \sN, 2\leq d\leq D$), which is formulated as:
\begin{equation}
\label{eq:ld}
    [\mE^d || \mZ^d] = L^d ([\mE^{d-1} || \mZ^{d-1}]) \in \R^{(V+P)\times F}.
\end{equation}
With the modulated node representations, we can get the predicted label for any node $v_j$ by applying a simple classifier, $f_\psi$ (e.g. SVM, Logistic Regression, shallow MLP, etc.):
\begin{equation}
    y = f_\psi(\ve_j^D). 
\end{equation}
Then, for each target $N$-way $K$-shot FSNC task $\mathcal{T}_i = \{\sS_i, \sQ_i\}$, we can predict labels for all the few labeled nodes in the support set $\sS_i$, and calculate the Cross-entropy loss, $\mathcal{L}_{CE}$, to update the prompt parameters $\mP$ and the simple classifier $f_\psi$. This optimization procedure can be formulated as:
\begin{equation}
    \label{eq:ft}
    \mP,\psi = \arg\min_{\mP,\psi} \mathcal{L}_{CE}(\sS_i;\mP,\psi)
\end{equation}
Finally, following the same procedure, we use the fine-tuned prompt $\mP$, classifier $f_\psi$, and node representations from the pretrained GT to predict labels for unlabeled nodes in the query set $\sQ_i$. It is notable that the parameters of the pretrained GT are frozen throughout the VNT process and are recycled for all downstream FSNC tasks. In other words, to adapt to a new FSNC task, we only need to train a small prompt tensor $\mP$ to modulate the intermediate-layer node representations to be customized by the few labeled nodes, which is \textbf{computationally similar to training a very shallow MLP}. This signifies that the proposed VNT method requires a low per-task storage and computation cost to retain its effectiveness. 

According to our experiments, the proposed VNT method can still achieve competitive performance even if no label in base classes is used. Furthermore, since during each downstream FSNC task, those virtual node prompts are tuned via conditioning on the same frozen GT backbone, we interpret the learned task prompts as task embeddings to form a semantic space of FSNC tasks. We give explanations in Section~\ref{exp:link}. Based on this interpretation, we propose a novel \textit{Graph-based Prompt Transfer} method to tackle scenarios where labeled nodes are available in the base classes.

\subsection{Prompt Transferring via Graph-based Pseudo Prompt Evolution}
For real-world scenarios, there could exist sparsely labeled nodes in base classes, i.e., a few source tasks exist, or $M\neq0$. On the other hand, we find that first training a prompt on one FSNC task, and then using the resulting prompt as the initialization for the prompt of another task could outperform tuning the virtual node prompt from scratch. We give a motivating example in Section~\ref{exp:link}. Inspired by this phenomenon, in this section, we propose a novel \textit{Graph-based Pseudo Prompt Evolution} (GPPE) mechanism that transfers the generalizable knowledge within tasks from base classes to target downstream FSNC tasks. The motivation behind prompt transferring is that, since all the source tasks and target tasks are sampled from the same input graph $\mathcal{G}$, incorporating context from all the individual source tasks will likely yield positive transfer. The details of GPPE are as follows.

To start with, as discussed in Section~\ref{sec:problem}, we assume that, for all the nodes in base classes, there exists a subset of nodes that can form $M$ $N$-way $K$-shot $R$-query node classification tasks, termed as \textit{source tasks}. Note that $M,N,K,R \ll~|V|$, so the labels can be very sparse and the value of $M$ determines the label sparsity in the base classes. In practice, we achieve this by random sampling $M$ FSNC tasks from base classes. Without more explanation, we use $M=48$, $R=10$ as default for experiments. Next, following the same procedure in Section~\ref{sec:NPT}, we first pretrain the GT encoder on the whole graph in an unsupervised manner, then we add virtual node prompts for each source task, and finally, those prompts are tuned on such $M$ source tasks. In this way, we obtain $M$ prompts, each of which can be interpreted as a \textit{task embedding}. Notably, according to~\eqref{eq:ft}, while performing VNT on the $M$ source tasks, we only use the support set to optimize the prompt. These $M$ prompts are stored as a \textit{prompt dictionary} $\mathcal{D} = [\mP_1^\prime;...;\mP_m^\prime;...;\mP_M^\prime] \in \sR^{M\times P \times F}$ for transferring knowledge to target FSNC tasks. The required space to store this dictionary is $\mathcal{O}(M\cdot P \cdot F)$, and as $M,P,F$ are small constants, storing this dictionary will not take much extra space, compared to the storage for the weights of the GT or the node embeddings. To further reinforce positive transfer from source tasks to target tasks, we propose a \textit{Prompt Evolution} module to refine those learned representations of virtual node prompts on each target FSNC task based on the task embeddings of all source FSNC tasks. We propose to use a fully connected Graph Attention Network (GAT)~\cite{Velickovic:2018we} to model the relations between these prompts and propagate context knowledge from all source FSNC tasks, where all the task-specific prompts can be regarded as the nodes in the GAT model. We choose GAT as the prompt evolution module for its desired properties: GAT can be inductive and is permutation invariant to the order of learning from source tasks. 

For training the prompt evolution module, we draw inspiration from meta-learning~\cite{finn2017model}, where a small classification task is constructed episodically to mimic the test scenarios and enable learning on the meta-level beyond a specific task. Similarly, in each episode, we randomly select one task $\mathcal{T}_m^\prime = \{\sS_m^\prime, \sQ_m^\prime\}$ as a pseudo target FSNC task, and the rest are still regarded as source tasks to extract transferable knowledge. The prompt evolution module will be trained through a number of episodes till its convergence. As the episodes iterate through the prompt dictionary $\mathcal{D}$, the prompt evolution module learns to refine all the prompt representation within $\mathcal{D}$, and simultaneously, learns to adapt to a target FSNC task given a set of source tasks.
Here, we illustrate the detailed learning procedure for one episode. We first compute a relation coefficient $c_{m,k}$ between the prompt of the $m$-th pseudo target FSNC task $\mathcal{T}_m^\prime$ and the $k$-th prompts for a source task in the dictionary $\mathcal{D}$. The coefficient is calculated based on the following kernel function:
\begin{equation}
    c_{m,k} = \langle  \bm\theta(\mP_m^\prime) , \bm\theta(\mP_k^\prime) \rangle,
\end{equation}
where $\bm{\theta}$ is a shallow MLP that projects the original prompts to a new metric space. Let $\langle \cdot \; , \; \cdot \rangle$ denote a similarity function. In this paper, we use cosine similarity for its simplicity and effectiveness. We then normalize all the coefficients with the softmax function
to get the final attention weights corresponding to the prompt  $\mP_m^\prime$ of the current pseudo target FSNC task:
\begin{equation}
    a_{m,k} = \frac{\exp{(c_{m,k})}}{\exp(\sum_{h=1}^{|\mathcal{D}|}c_{m,h})}.
\end{equation}
Based on the learned coefficients, the GAT model aggregates information from all the prompts learned from the source tasks in the graph and fuses it with $\mP_m^\prime$, the original prompt representation from a pseudo target task, to obtain a refined prompt $\Tilde{\mP}_m^\prime$:
\begin{equation}
    \Tilde{\mP}_m^\prime = {\mP}_m^\prime + (\sum^{|\mathcal{D}|}_{k=1} a_{m,k} \mL {\mP}_k^\prime),
\end{equation}
where $\mL$ denotes the weight matrix of a linear transformation. Then, with the refined prompt representation $\Tilde{\mP}_m^\prime$ and the simple classifier $f_\psi$, we use it to predict the labels of nodes in the corresponding query set $\sQ_m^{\prime}$. Based on the predicted labels, Cross-entropy loss is adopted to update the prompt evolution module:
\begin{equation}
    \bm\theta, \mL, \psi = \arg \min_{\bm\theta, \mL, \psi} \mathcal{L}_{CE} (\sQ_m^{\prime}, \Tilde{\mP}_m^\prime; \bm\theta, \mL, \psi).
\end{equation}
Once the prompt evolution module is learned, we freeze the parameters in the module and deploy it for any target FSNC task $\mathcal{T}_i = \{\sS_i, \sQ_i\}$ to get the refined prompt representation $\Tilde{\mP}$. Next, we train a separate simple classifier $f_{\psi}$ for final predictions:
\begin{equation}
    \label{eq:ft2}
    \psi = \arg\min_{\psi} \mathcal{L}_{CE}(\sS_i,\Tilde{\mP};\psi).
\end{equation}
The proposed GPPE module also has a similar parameter number as a shallow MLP, and we find that the proposed GPPE can be learned even if $M$, the number of source tasks, is very small. We give the analysis on the effect of $M$ in Fig.~\ref{fig:m} in Section~\ref{app:m}. The results show that GPPE improves the performance of VNT even with a very small number of source FSNC tasks (e.g., $M=8$). The process of our framework is illustrated in Fig.~\ref{fig:framework}.

\section{Experimental Study}
\label{sec:exp}
\subsection{Experimental Settings}
 
We conduct systematic experiments to compare the proposed VNT method with the baselines on the few-shot node classification task. In this work, we consider two categories of baselines, i.e., \textit{meta-learning} based methods, graph contrastive learning (GCL) based \textit{Transductive Linear Probing} (TLP) methods~\citep{tan2022transductive}, and \textit{prompting methods} on graphs. For meta-learning, we test typical methods (fully-supervised) including: \textbf{ProtoNet}~\cite{snell2017prototypical}, \textbf{MAML}~\cite{finn2017model}, \textbf{Meta-GNN}~\citep{zhou2019meta}, \textbf{G-Meta}~\citep{huang2020graph},
\textbf{GPN}~\citep{ding2020graph},
\textbf{AMM-GNN}~\citep{wang21AMM}, and \textbf{TENT}~\citep{wang2022task}. For GCL-based TLP methods, we evaluate TLP with self-supervised GCL methods including: \textbf{MVGRL}~\citep{hassani2020contrastive}, \textbf{GraphCL}~\citep{you2020graph}, \textbf{GRACE}~\citep{zhu2020deep}, \textbf{BGRL}~\citep{thakoor2021bootstrapped}, \textbf{MERIT}~\cite{jin2021multi}, and \textbf{SUGRL}~\cite{mo2022simple}. For {prompting methods} on graphs, we evaluate \textbf{GPPT}~\citep{sun2022gppt} and \textbf{Graph Prompt}~\citep{liu2023graphprompt}. For those GCL-based methods and the proposed VNT, we choose Logistic Regression as the classifier $f_\psi$. 
For comprehensive studies, we report the results of those methods on four prevalent real-world graph datasets:  \texttt{CoraFull}~\citep{bojchevski2018deep},  \texttt{ogbn-arxiv}~\citep{hu2020open}, \texttt{Cora}~\citep{yang2016revisiting}, \texttt{CiteSeer}~\citep{yang2016revisiting}. Each dataset is a graph that contains a considerable number of nodes. This ensures that the evaluation involves various tasks for a more comprehensive evaluation. A detailed description of those datasets is provided in Appendix~\ref{app:datasets}, with their statistics and class splits in Table~\ref{tab:statistics} in Appendix~\ref{app:stat}. For explicit comparison, we compare our method with all the baselines under various $N$-way $K$-shot $10$-query settings. The default values of the dictionary size $M$ and the query set size $R$ are $48$ and $10$, respectively. 

\vspace{-0.1cm}
\subsection{Comparable Study}
\begin{table*}[htbp]
\caption{\label{tab:all_result}The overall comparison between the proposed VNT method and meta-learning or self-supervised GCL-based TLP methods under different settings. Accuracy ($\uparrow$) and confident interval ($\downarrow$) are in~$\%$. The best results are \textbf{bold}, and the second best results in each category of methods are \underline{underlined}. OOM denotes the out-of-memory issue. 
}
\vspace{-0.1cm}
\scalebox{0.87}{
\begin{tabular}{@{}ccccccccc@{}}
\toprule
\textbf{Dataset}                         & \multicolumn{2}{c}{\texttt{CoraFull}} & \multicolumn{2}{c}{\texttt{Ogbn-arxiv}} & \multicolumn{2}{c}{\texttt{CiteSeer}} & \multicolumn{2}{c}{\texttt{Cora}}    \\ \midrule
\textbf{Setting}                         & 5-way 1-shot  & 5-way 5-shot & 5-way 1-shot   & 5-way 5-shot  & 2-way 1-shot  & 2-way 5-shot & 2-way 1-shot & 2-way 5-shot \\ \midrule
MAML                        & 22.63$\pm$1.19&27.21$\pm$1.32&27.36$\pm$1.48&29.09$\pm$1.62&52.39$\pm$2.20&54.13$\pm$2.18   & 53.13$\pm$2.26   & 57.39$\pm$2.23   \\
ProtoNet                        & 32.43$\pm$1.61&51.54$\pm$1.68&37.30$\pm$2.00&53.31$\pm$1.71&52.51$\pm$2.44&55.69$\pm$2.27   & 53.04$\pm$2.36   & 57.92$\pm$2.34   \\
Meta-GNN                        & 55.33$\pm$2.43    & 70.50$\pm$2.02   & 27.14$\pm$1.94     & 31.52$\pm$1.71    & 56.14$\pm$2.62    & 67.34$\pm$2.10   & 65.27$\pm$2.93   & 72.51$\pm$1.91   \\
GPN                             & 52.75$\pm$2.32    & 72.82$\pm$1.88   & 37.81$\pm$2.34     & 50.50$\pm$2.13    & 53.10$\pm$2.39    & 63.09$\pm$2.50   & 62.61$\pm$2.71   & 67.39$\pm$2.33   \\
AMM-GNN & 58.77$\pm$2.32    & 75.61$\pm$1.78   & 33.92$\pm$1.80     & 48.94$\pm$1.87    & 54.53$\pm$2.51    & 62.93$\pm$2.42   & 65.23$\pm$2.67   & \underline{82.30$\pm$2.07}   \\
G-Meta                          & \underline{60.44$\pm$2.48}    & \underline{75.84$\pm$1.70}   & 31.48$\pm$1.70     & 47.16$\pm$1.73    & 55.15$\pm$2.68    & 64.53$\pm$2.35   & \underline{67.03$\pm$3.22}   & 80.05$\pm$1.98   \\
TENT                            & 55.44$\pm$2.08    & 70.10$\pm$1.73   & \underline{48.26$\pm$1.73}     & \underline{61.38$\pm$1.72}    & \underline{62.75$\pm$3.23}    & \underline{72.95$\pm$2.13}   & 53.05$\pm$2.78   & 62.15$\pm$2.13   \\ \midrule
MVGRL                           & 59.91$\pm$2.39    & 76.76$\pm$1.63   & OOM            & OOM           & 64.45$\pm$2.77    & 80.25$\pm$1.82   & 71.17$\pm$3.04   & 89.91$\pm$1.44   \\
GraphCL                         & 64.20$\pm$2.56    & 83.74$\pm$1.46   & OOM            & OOM           & {73.51$\pm$3.09}    & \underline{92.38$\pm$1.24}   & {73.50$\pm$3.18}   & {92.35$\pm$1.30}   \\
GRACE                           & {66.69$\pm$2.26}    & {84.06$\pm$1.43}   & OOM            & OOM           & 69.85$\pm$2.75    & 85.93$\pm$1.57   & 69.13$\pm$2.69   & 88.68$\pm$1.37   \\
BGRL                            & 43.83$\pm$2.11    & 70.44$\pm$1.62   & {36.76$\pm$1.74}     & {53.44$\pm$0.36}    & 54.32$\pm$1.63    & 70.50$\pm$2.11   & 60.14$\pm$2.33   & 79.86$\pm$1.92   \\
MERIT                            & {73.38$\pm$2.25}&\underline{87.66$\pm$1.43}&OOM&OOM&{64.53$\pm$2.81}&{90.32$\pm$1.66}   & 67.67$\pm$2.99   & \underline{95.42$\pm$1.21}   \\
SUGRL                            & \textbf{77.35$\pm$2.20}&{83.96$\pm$1.52}&\underline{60.04$\pm$2.11}&\underline{77.52$\pm$1.45}&\textbf{77.34$\pm$2.83}&86.32$\pm$1.57 & \underline{82.35$\pm$2.20} & 92.22$\pm$1.15   \\\midrule
GPPT                          & {62.35$\pm$2.34}    & {73.68$\pm$2.24}   & {40.36$\pm$1.68}     & {51.68$\pm$1.92}    & {68.93$\pm$2.20}    & {82.53$\pm$1.86}   & {70.32$\pm$1.86}   & {85.58$\pm$1.73}   \\
Graph Prompt                         & {72.45$\pm$2.08}    & {81.29$\pm$2.36}   & {44.58$\pm$1.84}     & {75.62$\pm$1.96}    & {69.85$\pm$2.26}    & {85.26$\pm$1.78}   & {78.65$\pm$1.98}   & {89.38$\pm$1.96}   \\
VNT (\textbf{Ours.})                          & {68.50$\pm$2.13}    & {84.56$\pm$2.15}   & {50.40$\pm$1.97}     & {74.91$\pm$1.87}    & {70.60$\pm$2.15}    & {86.23$\pm$1.75}   & {84.50$\pm$1.94}   & {90.50$\pm$1.55}   \\
VNT-GPPE (\textbf{Ours.})                          & \underline{76.68$\pm$2.25}    & \textbf{88.75$\pm$2.07}   & \textbf{61.34$\pm$1.86}     & \textbf{79.93$\pm$1.69}    & \underline{75.85$\pm$2.45}    & \textbf{93.46$\pm$1.72}   & \textbf{88.62$\pm$2.12}   & \textbf{95.65$\pm$1.51}   \\\bottomrule
\end{tabular}}
\vspace{-0.15cm}
\end{table*}

Table~\ref{tab:all_result} presents the performance comparison of all the methods on the few-shot node classification task. Specifically, we present results under four different few-shot settings for a more comprehensive comparison: 5-way 1-shot, 5-way 5-shot, 2-way 1-shot, and 2-way 5-shot. We choose the average classification accuracy and the $95\%$ confidence interval over 5 repetitions with different random seeds as the evaluation metrics. For each repetition, we sample 100 FSNC tasks for evaluation and calculate the evaluation metrics. From Table \ref{tab:all_result}, we obtain the following observations:
\setlength\itemsep{0.1em}
\begin{itemize}
    \item Generally speaking, for both TLP and the proposed GT, \textbf{self-supervised pretraining can outperform the meta-learning-based method}. However, one most recent pretraining method, BGRL, when transferred for downstream FSNC tasks, shows surprisingly frustrating performance. This further validates \textbf{the impact from the gap of training objective} between pretexts and target FSNC tasks. The pretext of BGRL minimizes the \textit{Mean Square Error} of the original node representation and its slightly perturbed counterpart but does not enforce the model to discriminate between different nodes as the other GCL baselines do. The objective of this pretext deviates more from the downstream FSNC tasks, thus leading to worse results. We further show the impact of this in ablation studies in Section \ref{exp:ablation}.

    \item Even \textbf{without any label information from base classes}, the proposed method, VNT, outperforms meta-learning-based methods and most GCL-based TLP baselines. This demonstrates the superiority of the proposed VNT in terms of accuracy. The pretrained GT has learned substantial prior knowledge and the injected virtual node prompts effectively elicit the knowledge for different downstream FSNC tasks.

    \item Given \textbf{a few source FSNC tasks}, the proposed method, VNT-GPPE, consistently outperforms all the baselines, including existing prompt methods for graphs. This further validates that GPPE effectively generalizes the knowledge learned from the few source tasks to target tasks, yielding positive transfer.
    

    \item Compared to all the baselines, the proposed VNT-GPPE method is \textbf{more robust to extremely scarce label scenarios}, i.e, the number of labeled nodes in the support set $K$ equals 1. {The performance degradation resulting from decreasing the number of shots} $K$ is significant for all the methods.
    Smaller $K$ makes the encoder or the classifier more prone to overfitting, thus leading to worse generalization to query sets. In contrast, the proposed method injects virtual nodes into the model, which have separate learnable embeddings for different FSNC tasks. This implies that our method implicitly performs adaptable data augmentation for the few labeled nodes, which makes our framework more robust to tasks with extremely scarce labeled nodes. Also, the \textbf{improvement of involving GPPE is more significant when $K$ is extremely small}. This shows that transferring knowledge from source tasks helps to mitigate the overfitting on novel tasks. Further analysis and explanation are given in Section~\ref{exp:link}.
\end{itemize}

\vspace{-0.2cm}
\subsection{Ablation Study}
\label{exp:ablation}
\begin{table*}[htbp]
\caption{\label{tab:ablation}Ablation study on \texttt{Cora} and \texttt{Ogbn-arxiv} datasets to analyze the effectiveness of different components in our method.}
\centering
\scalebox{0.85}{
\begin{tabular}{@{}cccc|cc|cccc@{}}
\toprule
\multirow{2}{*}{\textbf{Encoder}} & \multirow{2}{*}{\textbf{Frozen}} & \multirow{2}{*}{\textbf{VNT}} & \multirow{2}{*}{\textbf{GPPE}} & \multicolumn{2}{c|}{\texttt{Cora}}                     & \multicolumn{4}{c}{\texttt{Ogbn-arxiv}}                                                                \\ \cmidrule(l){5-10} 
                                  &                                  &                                  &                                & 2-way 1-shot          & 2-way 5-shot          & 2-way 1-shot          & 2-way 5-shot          & 5-way 1-shot          & 5-way 5-shot          \\ \midrule
GCN                               &                                  &                                  &                                & 52.12$\pm$2.62          & 57.93$\pm$2.23          & 57.62$\pm$2.31          & 64.11$\pm$2.65          & 26.68$\pm$1.57          & 27.90$\pm$1.45          \\
GCN                               & $\checkmark$                     &                                  &                                & 68.43$\pm$2.94          & 78.20$\pm$2.83          & 65.21$\pm$2.86          & 77.10$\pm$2.46          & 38.47$\pm$1.77          & 51.46$\pm$1.69          \\ \midrule
GT                                &                                  &                                  &                                & 67.50$\pm$2.24          & 79.42$\pm$1.89          & 63.00$\pm$2.35          & 79.84$\pm$1.98          & 40.73$\pm$2.65          & 55.35$\pm$1.88          \\
GT                                & $\checkmark$                     &                                  &                                & 75.50$\pm$2.54          & 84.94$\pm$1.74          & 53.64$\pm$2.62          & 73.64$\pm$2.33          & 31.64$\pm$2.45          & 52.36$\pm$2.04          \\ \midrule
GT                                &                                  & $\checkmark$                     &                                & 77.85$\pm$1.99          & 85.43$\pm$1.84          & 71.82$\pm$2.58          & 82.73$\pm$2.14          & 36.36$\pm$2.74          & 65.45$\pm$2.31          \\
GT                                & $\checkmark$                     & $\checkmark$                     &                                & 84.50$\pm$1.94          & 90.50$\pm$1.55          & 82.00$\pm$1.77          & 87.27$\pm$1.64          & 50.40$\pm$1.97          & 74.91$\pm$1.87          \\ \midrule
GT                                & $\checkmark$                     & $\checkmark$                     & $\checkmark$                   & \textbf{88.62$\pm$2.12} & \textbf{95.65$\pm$1.51} & \textbf{85.35$\pm$1.72} & \textbf{89.98$\pm$1.66} & \textbf{61.34$\pm$1.86} & \textbf{79.93$\pm$1.69} \\ \bottomrule
\end{tabular}
}
\end{table*}

In this subsection, we conduct ablation studies to investigate the effectiveness of different components, i.e., VNT and GPPE, of the proposed framework. We consider both cases: when the encoder is frozen and when it is not frozen.
We present the results of experiments on the \texttt{Cora} and \texttt{Ogbn-arxiv} datasets, under different
$N$-way $K$-shot settings (similar results can be observed on the other datasets and settings). For the GCN baselines, following the common practice~\citep{zhou2019meta,ding2020graph}, we pretrain a 2-layer GCN using all the data from base classes with Cross-Entropy Loss. Specifically, Frozen means during fine-tuning, the GNN encoder is fixed, and only the classifier is fine-tuned. Prompt refers to the proposed virtual node tuning method. GPPE is the proposed graph-based pseudo prompt evolution module. Because GPPE is based on the proposed framework of VNT, for the tested variant with GPPE, we freeze the GT encoder and add virtual node prompts. The scores reported are averaged over 5 runs with different random seeds. The results of the ablation study are presented in Table~\ref{tab:ablation}, from which we draw the following conclusions:
\begin{itemize}
    \item Our simple implementation of \textbf{GT can consistently yield better results} than traditional GNNs, such as GCN. This is because the GT has a much larger number of parameters, making it capable of learning more complex relations among nodes. Besides, pretrained with the two pretext tasks, i.e., \textit{node attribute reconstruction} and \textit{structure recovery}, in a self-supervised manner, GT can learn more transferable graph patterns compared to those GCN-based methods.
    \item \textbf{Freezing the GNN encoder} during fine-tuning on the downstream FSNC tasks consistently leads to better results. This shows that fine-tuning the graph encoder on the few labeled nodes could lead to overfitting and negatively impact the quality of the learned node embeddings.
    \item The proposed method, VNT, which contains a frozen pretrained GT encoder with virtual node, exhibits \textbf{competitive performance} compared with vanilla GTs. This implies that the introduced virtual nodes can help the model modulate the learned substantial graph knowledge for each FSNC task while avoiding impairing the pretrained embeddings.
    \item The proposed method, VNT-GPPE, which involves the prompt evolution module achieves \textbf{the best performance}. This validates that the introduced GPPE module can effectively provide positive transfer from source tasks to target tasks and mitigate
the overfitting on novel tasks.
\end{itemize}



\subsection{Node Embedding Analysis}
\label{exp:clustering}
To explicitly illustrate the advantage of the proposed framework, in this subsection, we analyze the quality of the learned node representations from different training strategies. Particularly, we leverage two prevalent clustering evaluation metrics: \textit{Normalized
Mutual Information} (NMI) and \textit{adjusted random index} (ARI), on learned node embeddings clustered based on K-Means. Also, we deploy t-SNE to visualize them and compare them with those learned by baseline methods on the \texttt{CoraFull} dataset. We choose nodes from 5 randomly selected novel classes for visualization. The results are presented in Table~\ref{tab:cluster} and Fig.~\ref{fig:sne}. Complete results with all the baselines are included in Appendix~\ref{app:clustering}. We observe that the proposed VNT-GPPE method enhances the quality of the node representations of GTs and achieves the best clustering performance on novel classes. Also,  we find that a vanilla GT without VNT cannot learn node embeddings that are discriminative enough compared to strong baselines like TENT and SUGRL. However, when equipped with the proposed VNT, a GT can learn highly discriminative node embeddings. Furthermore, GPPE can significantly improve the performance of VNT. This also authenticates that the introduced prompts help to elicit more customized knowledge for each downstream FSNC task, and GPPE can effectively transfer the knowledge learned from source FSNC tasks to target FSNC tasks.

\begin{table}[htbp]
		\setlength\tabcolsep{9.5pt}
	\small
		\centering
		\renewcommand{\arraystretch}{1.2}
		\caption{\label{tab:cluster}The overall NMI ($\uparrow$) and ARI ($\uparrow$) scores of baselines and ablated variants of the proposed framework on \texttt{CoraFull} and \texttt{CiteSeer} datasets. 
  }
        \vspace{-0.05in}
        \scalebox{1.}{
		\begin{tabular}{c||c|c||c|c}
			\hline
			\textbf{Dataset}&\multicolumn{2}{c||}{\texttt{CoraFull}}&\multicolumn{2}{c}{\texttt{CiteSeer}}
			\\
			\hline
						\textbf{Metrics}&\multicolumn{1}{c|}{\textbf{NMI}}&\multicolumn{1}{c||}{\textbf{ARI}}&\multicolumn{1}{c|}{\textbf{NMI}}&\multicolumn{1}{c}{\textbf{ARI}}\\

					\hline
			TENT&$0.5760$&$0.4652$&$0.0930$&$0.0811$\\\hline

   
SUGRL&$0.7680$&$0.7049$&$0.3952$&$0.4460$\\\hline


\hline

GT&$0.5225$&$0.3864$&$0.3452$&$0.3189$\\\hline
VNT&${0.7768}$&${0.6427}$&${0.5998}$&${0.6331}$\\\hline
VNT-GPPE&$\mathbf{0.7927}$&$\mathbf{0.7075}$&$\mathbf{0.6324}$&$\mathbf{0.6762}$\\\hline

	\hline
				
\end{tabular}}
		\label{tab:nmi_result}
  \vspace{-0.55cm}
	\end{table}

\begin{figure}[htbp]
		\centering
	\subcaptionbox{TENT}{
\includegraphics[width=0.15\textwidth]{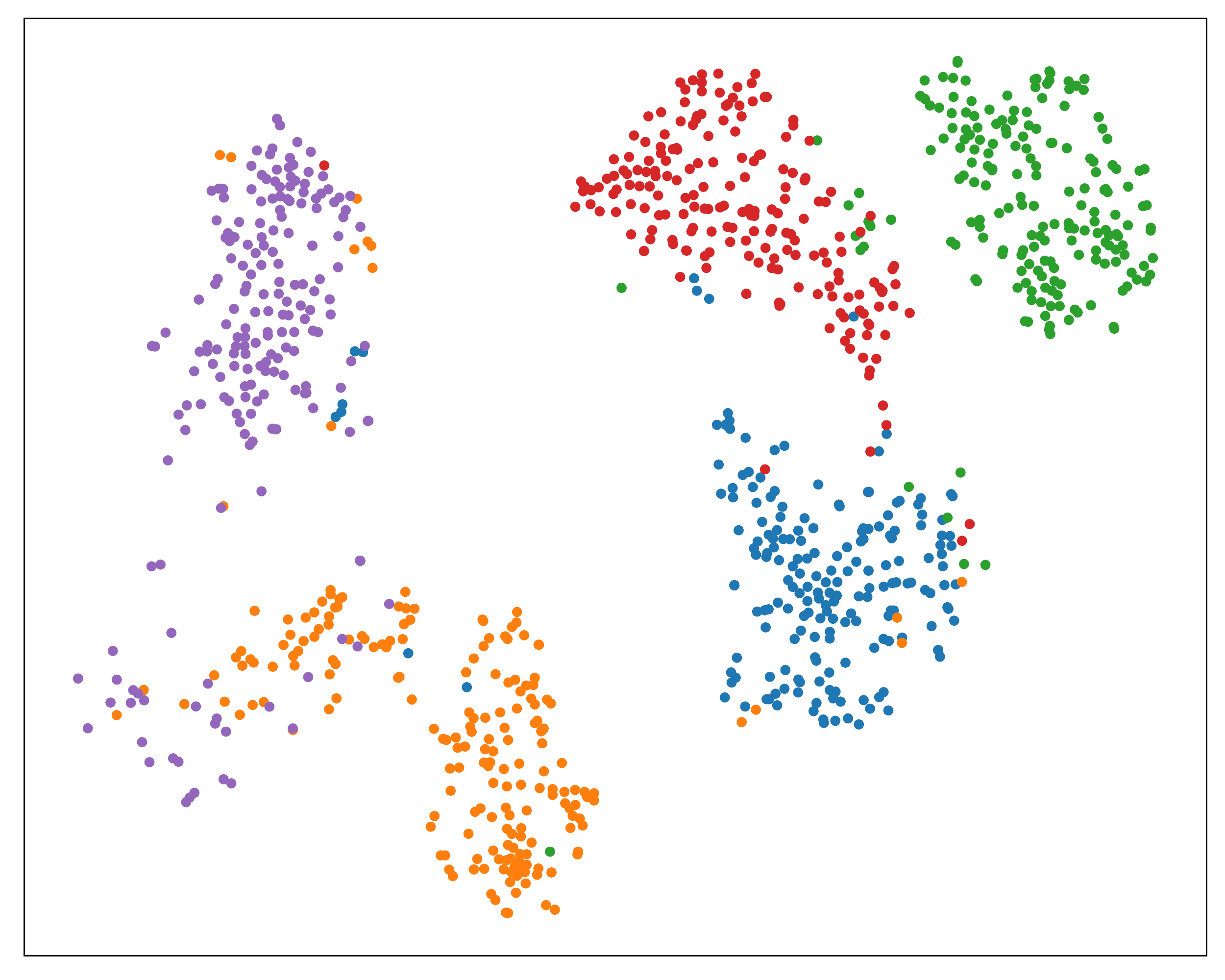}}
	\subcaptionbox{GRACE}{
\includegraphics[width=0.15\textwidth]{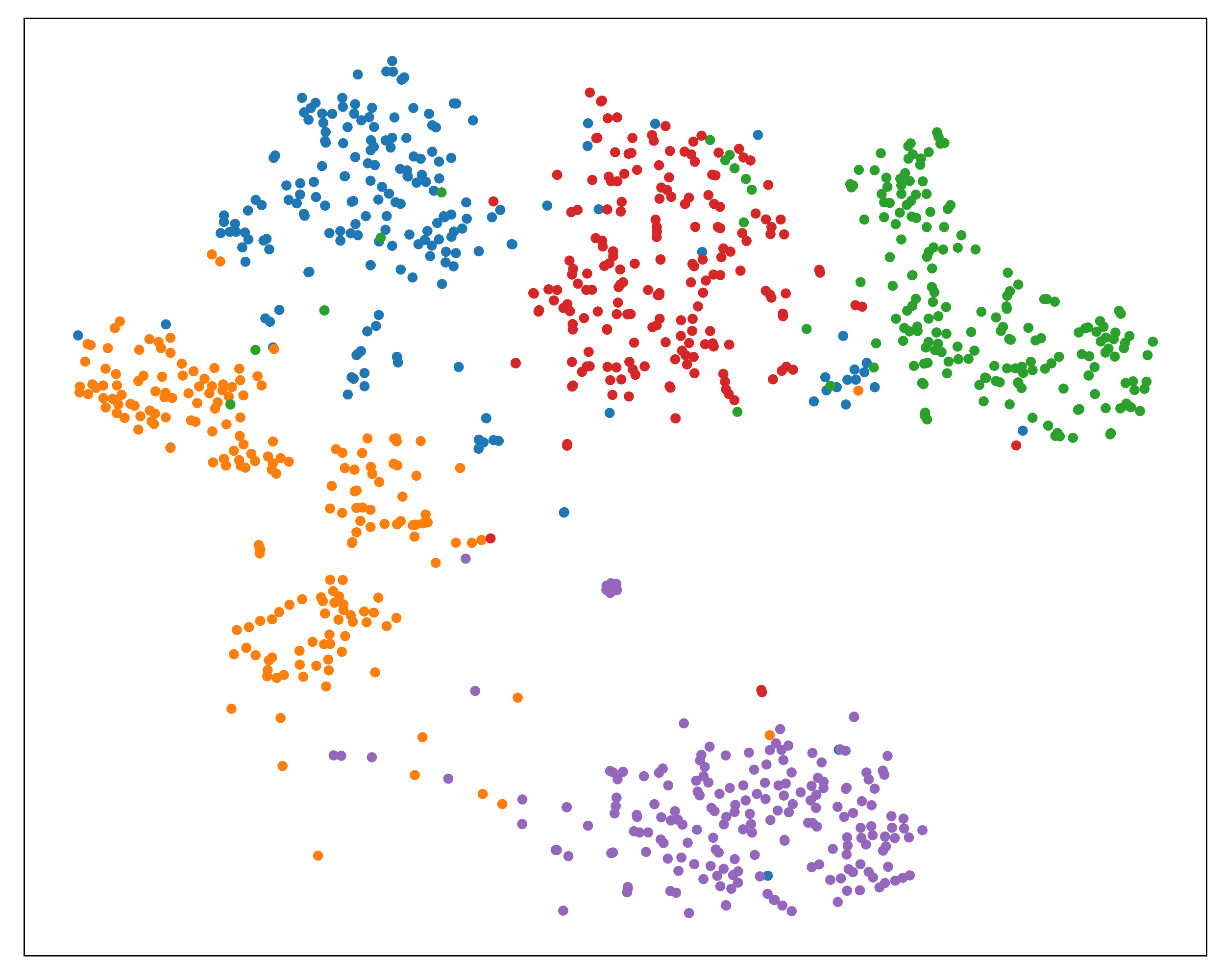}}
\subcaptionbox{SUGRL}{
\includegraphics[width=0.15\textwidth]{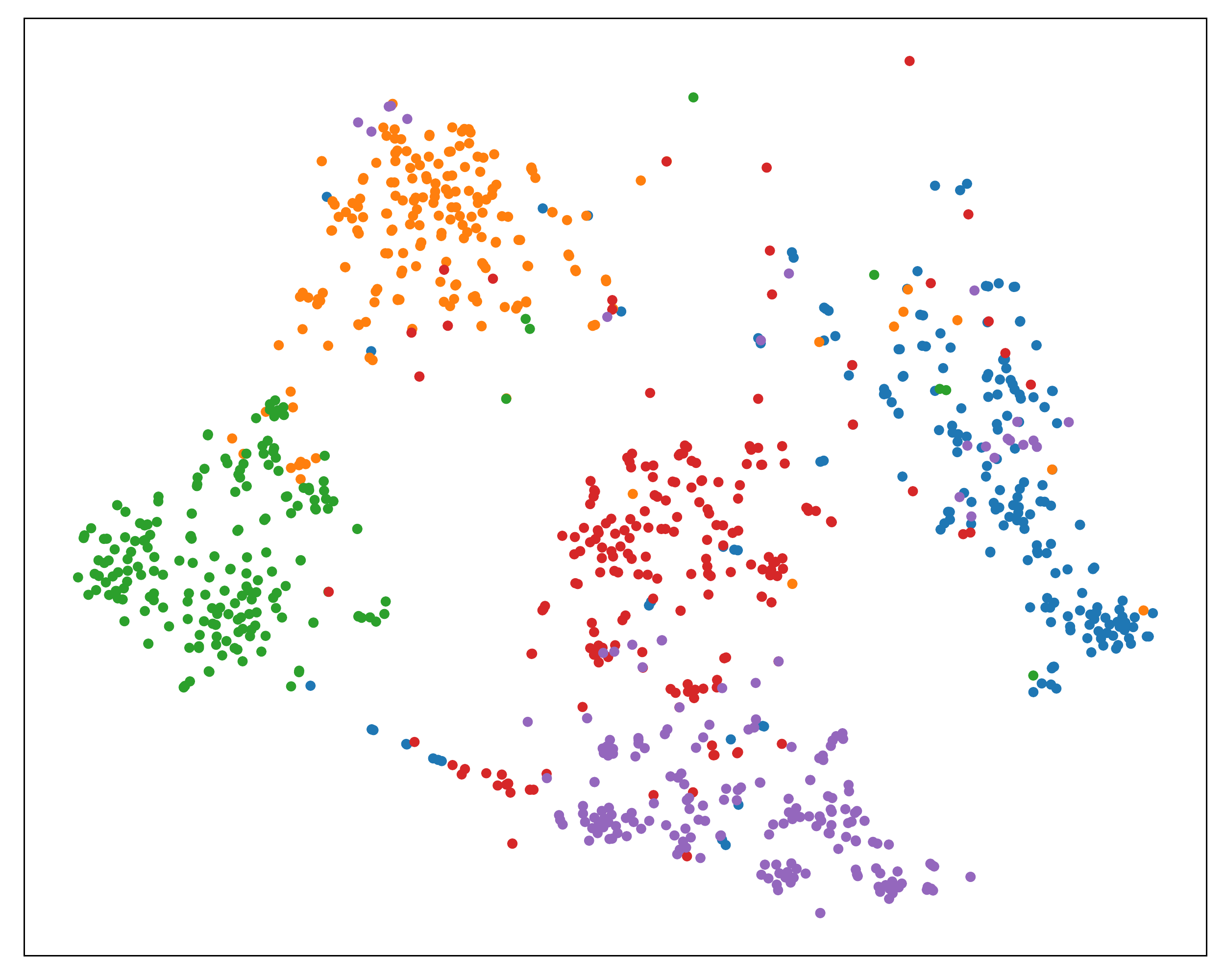}}
    \subcaptionbox{GT}{
\includegraphics[width=0.15\textwidth]{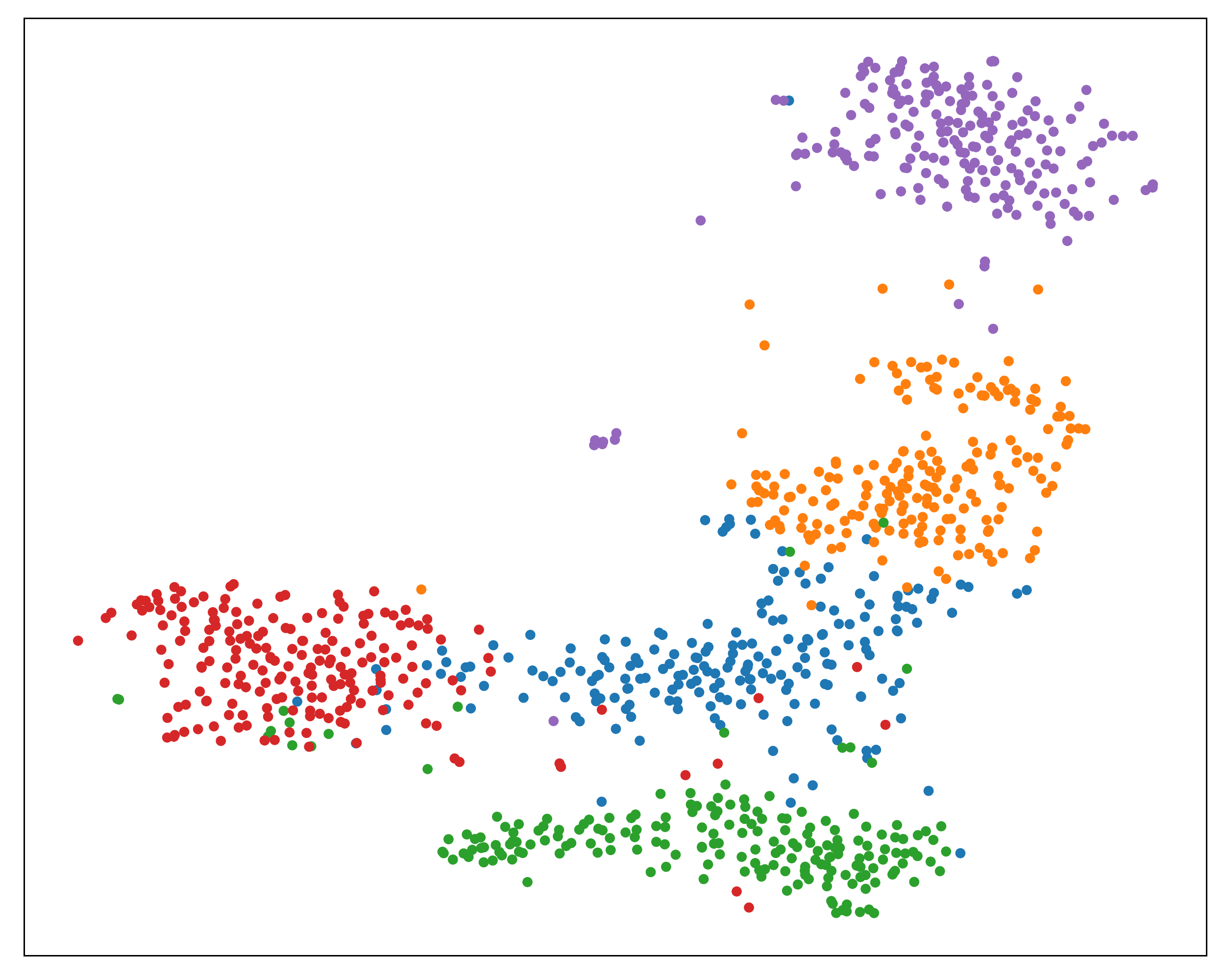}}
\subcaptionbox{VNT}{
\includegraphics[width=0.15\textwidth]{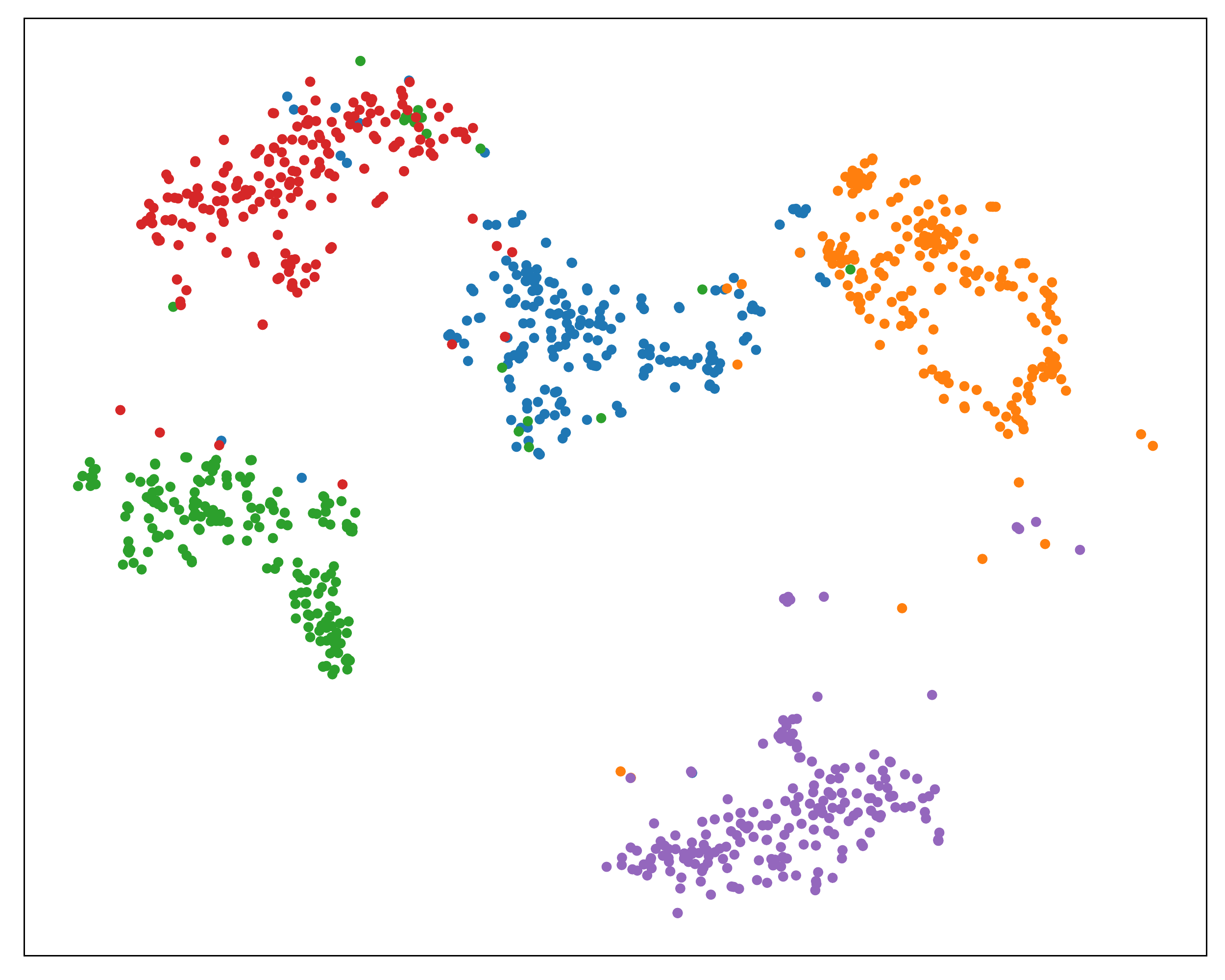}}
\subcaptionbox{VNT-GPPE}{
\includegraphics[width=0.15\textwidth]{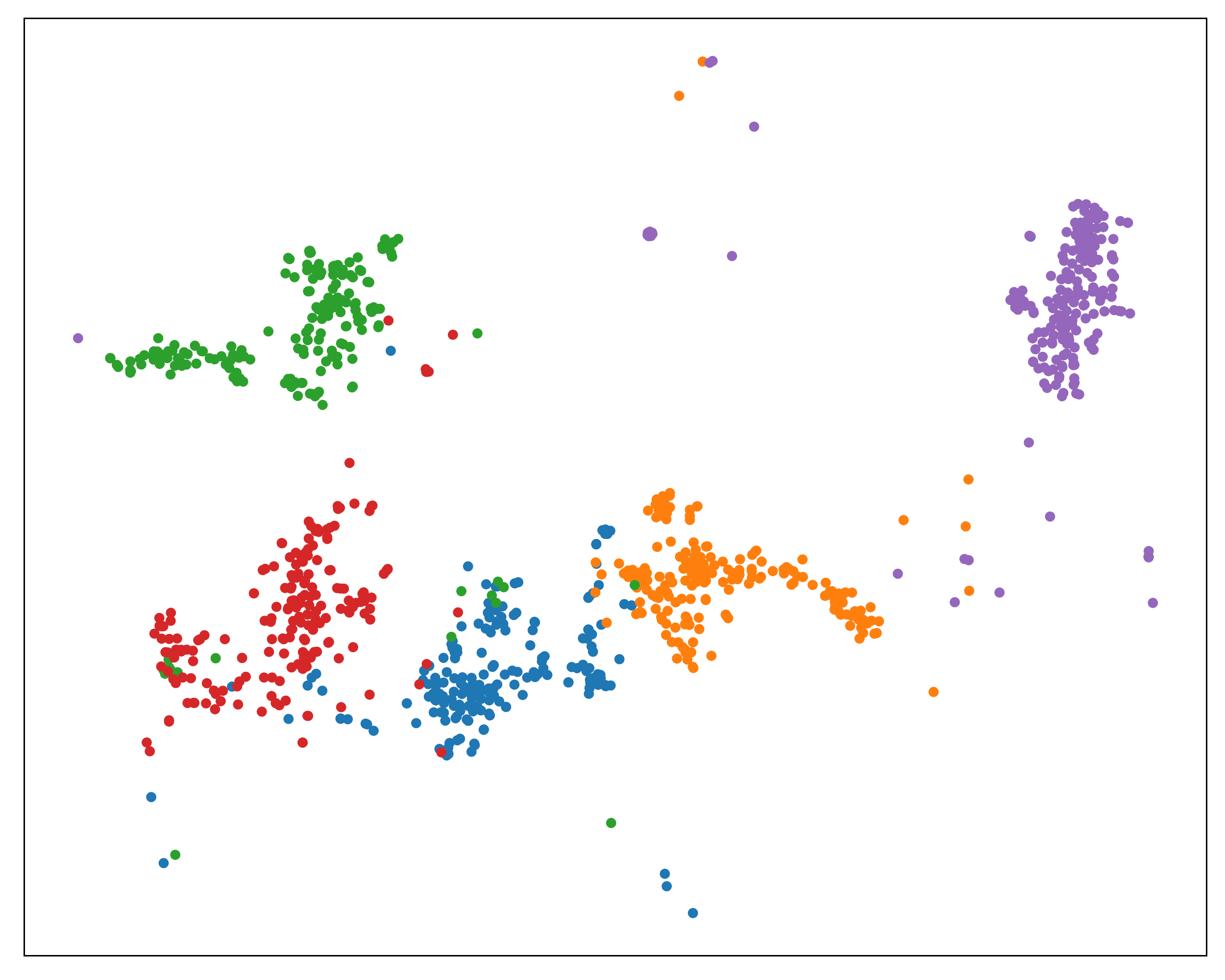}}

\vspace{-0.25cm}
\caption{\label{fig:sne}The t-SNE results on \texttt{CoraFull} ($5$-way $5$-shot).}
\vspace{-0.35cm}
	\end{figure}

\subsection{Design Discussion}

\subsubsection{Interpretation of Virtual Nodes as Prompt}
\label{exp:link}

\begin{figure}[htbp]
		\centering
\scalebox{0.55}{
\includegraphics[width=0.8\textwidth]{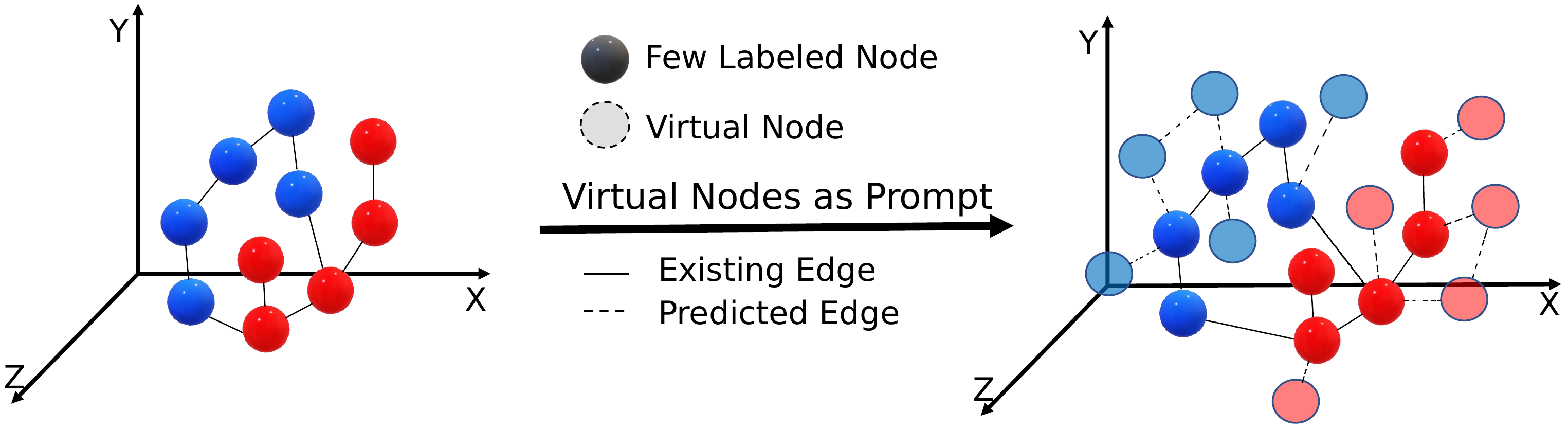}}
\vspace{-0.1cm}
\caption{The illustration of effect from introduced virtual nodes under $2$-way node classification setting. Different colors indicate different node classes.}
\label{fig:interp}
	\end{figure}

\begin{table}[htbp]
\vspace{-0.35cm}
\caption{\label{tab:interp}The $L_2$ distance and $cosine$ similarity scores between prompting virtual nodes and real nodes from two novel classes on \texttt{Cora} dataset. P denotes virtual node prompts, and N denotes existing nodes.}
\scalebox{1.}{
\centering
\begin{tabular}{ccccc}
\hline
\textbf{Metrics} & $L_2$  & $Cosine$ & $L_2$  & $Cosine$ \\ \hline
\textbf{Nodes}   & \multicolumn{2}{c}{P in class-1} & \multicolumn{2}{c}{P in class-2} \\ \hline
N in class-1     & 0.0622         & 0.7625          & 0.6725          & 0.2344          \\
N in class-2     & 0.6520         & 0.8627          & 0.0548         & 0.1165          \\ \hline
\end{tabular}}
\vspace{-0.2cm}
\end{table}
In this study, we explore the interpretability of virtual node prompts in graph data. Unlike previous works in NLP, where prompts are composed of human language that is easily interpretable by humans, virtual nodes in graph data are injected into the node embedding space, making it difficult to understand their effect. 
To gain insight into the behavior of virtual node prompts, we exploit Walkpooling~\citep{pan2021neural}, which is the state-of-the-art for link prediction, to predict the links that are the most likely to exist between the virtual nodes and the few labeled nodes from the support set of each task. We train the model on the whole graph dataset and use it to predict the most possible links between the virtual nodes and the existing ones. Specifically, we only consider potential links with at least one virtual node as their vertex. Under the $2$-way few-shot node classification setting, we initialize half of the virtual node prompts as the prototype vector of the first class, and the other half of the virtual node as the prototype vector of the second class. As indicated in Fig.~\ref{fig:interp}, after the convergence of virtual node tuning, we find that the vertices of the most possible links always connect existing nodes with virtual prompt nodes from the same classes. This implies that the introduced virtual node prompts can learn node representations semantically similar to those from the same class, and \textbf{thus help push node representations from the same classes closer}. To further validate this, on \texttt{Cora} dataset, we calculate the average $cosine$ similarities and $L_2$ distances (normalized by the longest distance of any pair of nodes) for virtual nodes and existing nodes from the two novel classes. As presented in Table~\ref{tab:interp}, we can see that the virtual nodes and existing nodes from the same classes have smaller $L_2$ distances and larger $cosine$ similarities. We provide further analyses of the effect from different numbers of virtual nodes in Fig.~\ref{fig:prompt_len} in Appendix~\ref{app:prompt_len}.

\subsubsection{Motivation of Virtual Node Prompt Transfer}
\label{exp:gppe} 
\begin{figure}[htbp]
		\centering
	\subcaptionbox{Learned prompts reuse}{
\includegraphics[width=0.23\textwidth]{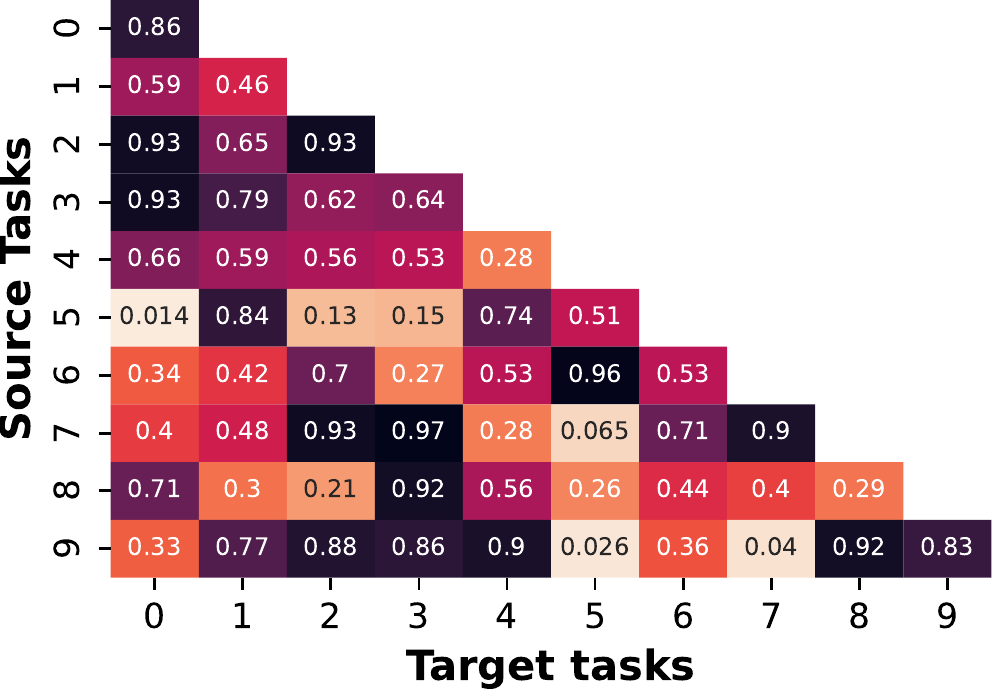}}
	\subcaptionbox{Learned prompts as initializer}{
\includegraphics[width=0.23\textwidth]{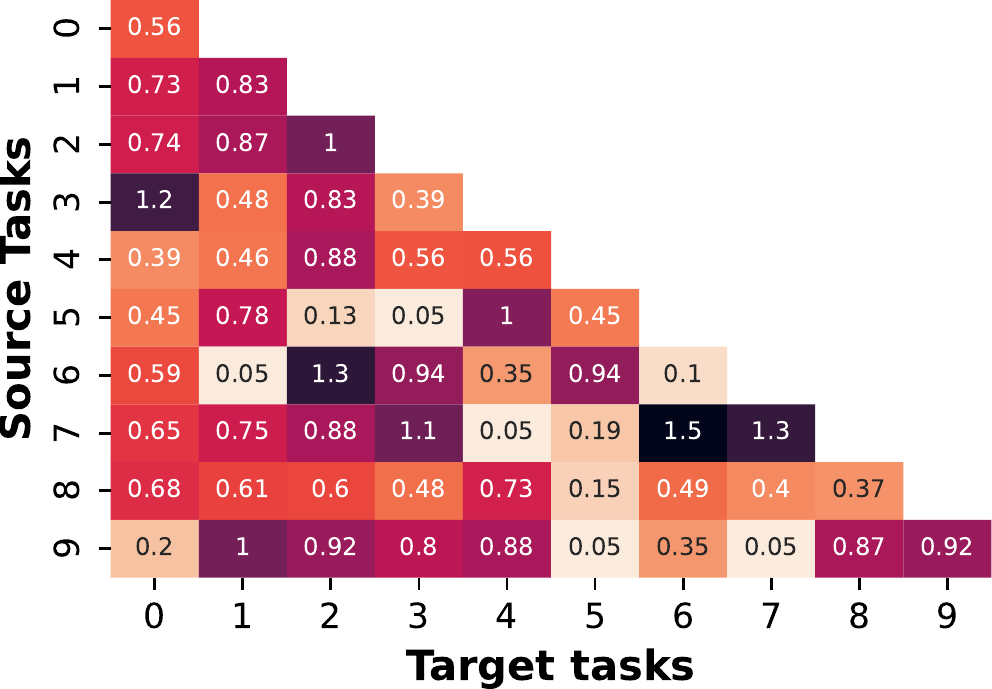}}
\vspace{-0.15cm}
\caption{\label{fig:gppe} Relative prompt transfer performance (transfer performance / original VNT performance) on the target tasks of the virtual node prompts trained on the source tasks.}
\vspace{-0.3cm}
\end{figure}

In this experiment, we empirically study the transferability across randomly sampled $10$ source FSNC tasks and $10$ target FSNC tasks from \texttt{CoraFull} dataset under the $5$-way $5$-shot setting. To test this, we first perform VNT on a source task and then directly reuse the learned virtual node prompts for other target tasks. As shown in Fig.~\ref{fig:gppe} (a), we observe that reusing the prompts learned from some source tasks will provide decent performance on corresponding target tasks. Then, we examine a very naive approach to transfer prompts: we use the learned virtual node prompt from a source task as the initializer of prompts for a target task, and then fine-tune it with VNT. As demonstrated in Fig.~\ref{fig:gppe} (b), we can see that through such a simple transfer, VNT can perform better on some target tasks than training from scratch. Both these results imply that selectively transferring learned knowledge from prompts learned in source tasks to target tasks will likely yield positive transfer. This experiment motivates us in designing the GPPE module.

\subsubsection{The Effect of Source Task Number ${M}$}
\label{app:m}
\begin{figure}[htbp]
\vspace{-0.05cm}
		\centering
\captionsetup[sub]{skip=-1pt}
\subcaptionbox{\texttt{CoraFull}}
{\includegraphics[width=0.235\textwidth]{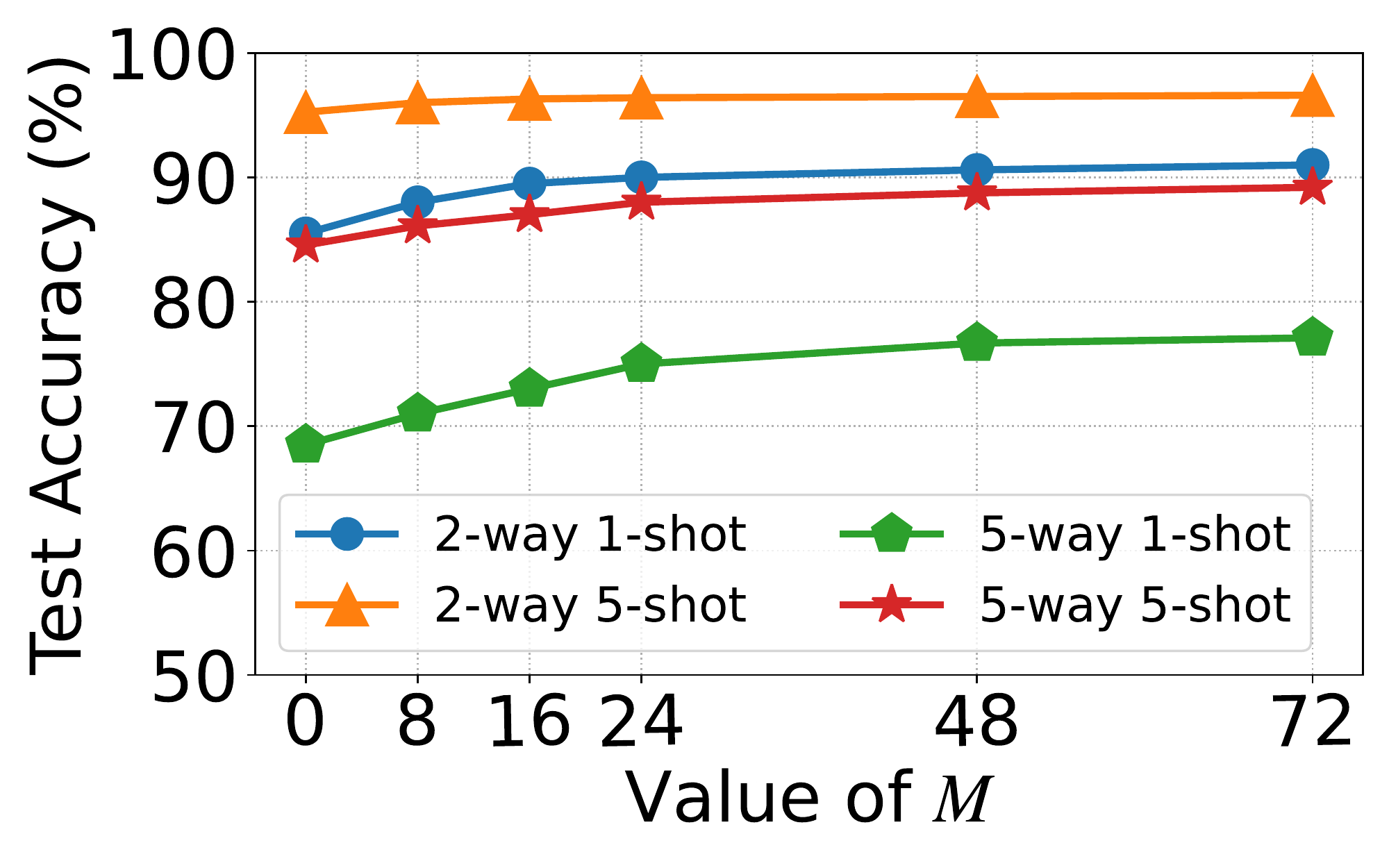}}
\subcaptionbox{\texttt{Ogbn-arxiv}}
{\includegraphics[width=0.235\textwidth]{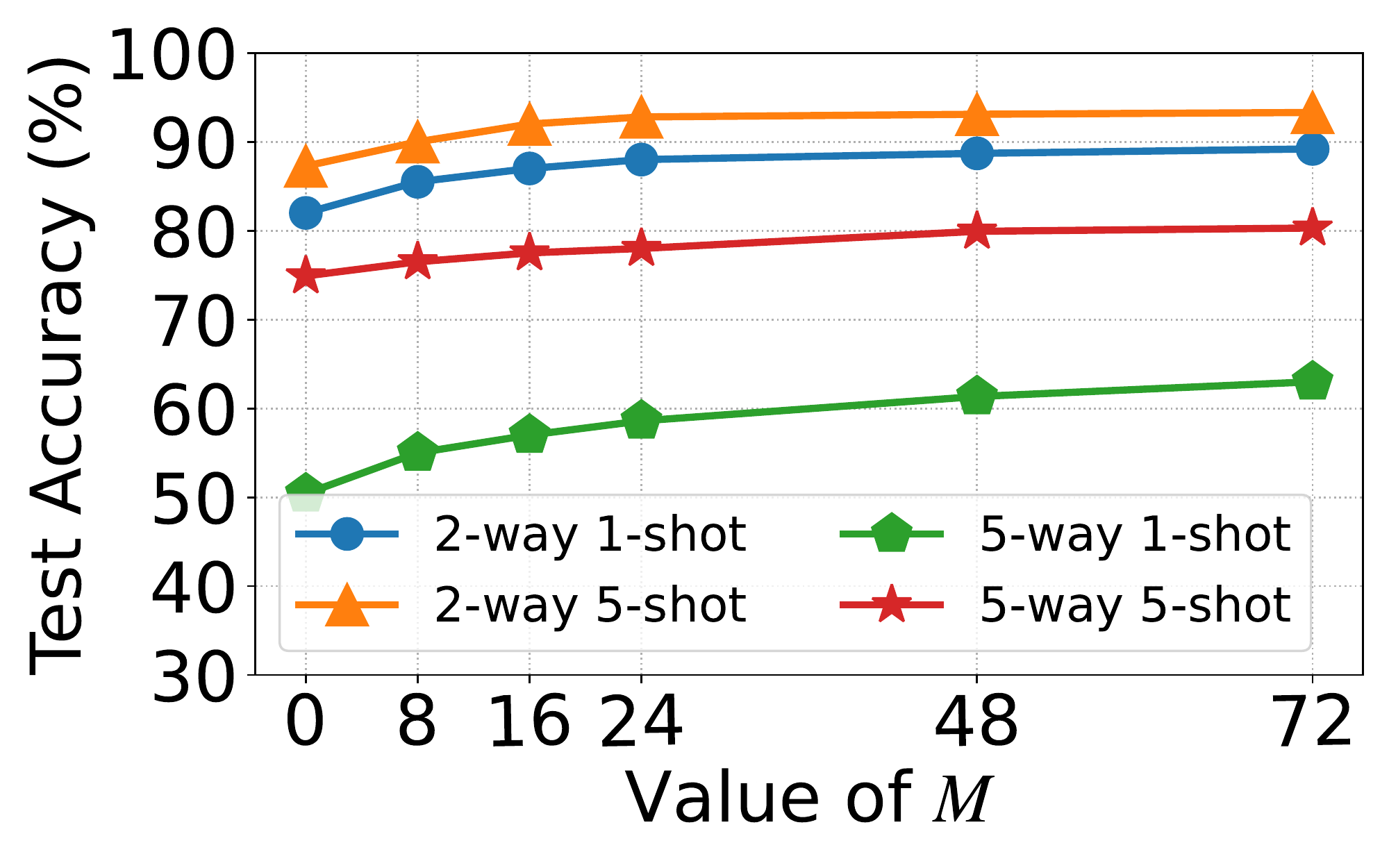}}
\vspace{-0.15in}
		\caption{The results of our proposed framework, VNT-GPPE, under different few-shot settings with varying values of the number of source FSNC tasks $M$ on CoraFull and Ogbn-arxiv.}
		\label{fig:m}
\vspace{-0.15in}
	\end{figure}

In this experiment, we evaluate the effect of the source task number $M$ on our framework, VNT-GPPE. A larger value of $M$ signifies more labeled nodes in base classes. $M=0$ means no source task exists. Thus, the framework will be reduced to VNT. Fig.~\ref{fig:m} reports the results of our framework with varying values of $M$ under different few-shot settings. From the results, we observe that generally increasing $M$ will lead to better performance. This is because more labeled nodes in base classes contain more transferable graph knowledge, and the proposed GPPE module can effectively transfer the learned knowledge to target novel classes. We choose $M=48$ as the default setting. An important observation is that, given a very small number of source FSNC tasks, e.g., $M=8$, GPPE can still improve VNT by a large margin. This shows that the proposed framework scales well to scenarios with sparsely labeled base classes.

\subsection{Sensitivity Analysis of VNT}
In this experiment, we aim to study the sensitivity of the VNT framework under various conditions. Specifically, we conduct experiments to evaluate its performance when no source tasks exist in the base classes, i.e., $M=0$. This setup allows us to observe the general performance of the VNT framework in different scenarios. The results of these experiments will provide valuable insights into the robustness and flexibility of the VNT framework, and help guide the design of future research in this area.
\subsubsection{Prompt Initialization}
\label{sec:init}
Noticeably, results in subsection~\ref{exp:link} imply that initializing the virtual nodes as prototype representations could benefit the virtual node tuning process. Intuitively, given a test node, an ideal model should produce an output node embedding that is close to the corresponding class prototype representations. To test this, we initialize equal portions of virtual prompt nodes to all novel node classes as their prototype representations. The results, presented in Table~\ref{tab:initial_ensemble} in Appendidx~\ref{app:design}, show that this simple initialization can consistently enhance the performance on downstream few-shot node classification tasks. This suggests that providing the model with hints about the target categories through initialization can help improve the optimization process.


\subsubsection{Prompt Ensemble}
Previous research by \citet{lester2021power} has demonstrated the efficiency of using prompts for ensembling, as the large transformer backbone can be frozen after pretraining, reducing the storage space required and allowing for efficient inference through the use of specially-designed batches~\citep{lester2021power,jia2022visual}.
Given such advantages, we investigate the effectiveness of enabling prompt ensembling for VNT. Concretely, to facilitate the ensembling with the prompt initialization strategy, we add independently sampled Gaussian noise tensors to 5 prompts for each FSNC task. Each prompt contains virtual nodes initialized as node class prototypes. We use majority voting to compute final predictions from the ensemble. Table~\ref{tab:initial_ensemble} in Appendidx~\ref{app:design} shows that the ensembled model outperforms the average or even the best single prompt counterpart.

\subsubsection{Effectiveness with regard to the Scale of GTs}\label{exp:scale}


In Fig.~\ref{fig:scale} in Appendix~\ref{app:design}, we present the accuracy of the proposed framework against the scale of the GT encoder as a heat map. We analyze the impact of varying the width (embedding size $F$) and the depth (number of transformer layers $D$) of the GT on the performance of our framework. The results shown are from the \texttt{Cora} dataset under the $2$-way $1$-shot setting and similar trends are observed on other datasets under different settings. We have the following findings:

\textbf{i.} Increasing the depth of the GT encoder does not necessarily improve the performance of downstream FSNC tasks. In fact, as the depth increases, the final accuracy may decrease. This is likely because the virtual node prompts are injected only before the first transformer layer, and their effect diminishes as they go deeper. This highlights the importance and effectiveness of the proposed VNT framework for better adaptation in FSNC tasks.
    
 \textbf{ii.} On the other hand, increasing the width of the GT encoder, i.e., the embedding size, leads to improved accuracy. A larger embedding size implies that the input node embeddings contain more detailed semantic and topological information, allowing the injected virtual nodes to modulate the node embeddings more precisely.




\section{Related Work}
\label{sec:backgroud}
\subsection{Few-shot Node Classification.}
Episodic meta-learning~\citep{finn2017model} has become the most dominant paradigm for Few-shot Node Classification (FSNC) tasks. It trains the GNN encoders by explicitly mimicking the test environment for few-shot learning~\citep{zhou2019meta,ding2020graph}.
Nonetheless, these methodologies depend on the supposition that 'base classes' are available, with ample labeled nodes per class for episode sampling. This leads to a limitation, as these existing techniques do not cater to the more general FSNC problem defined in our study.
\subsection{Graph Transformer.}
Graph Transformers (GTs) \citep{zhang2020graph,rong2020self,chen2022structure} are a new breed of GNNs that leverage the transformer architecture~\citep{vaswani2017attention}. GTs typically consist of two parts: an embedding network that projects raw graphs to the embedding space, and a transformer-based network that learns the complex relationships among the embeddings. Due to their vast number of parameters, GTs have the ability to capture a greater depth and complexity of knowledge. They are usually trained in a self-supervised way using pre-defined pretext tasks, such as such as node attribute reconstruction~\citep{zhang2020graph} and structure recovery~\citep{chen2022structure} to encode both topological and semantic information. Recent advancements in GTs aim to encode increasingly complex topological knowledge, such as adaptive mechanisms for position encoding from graph spectrums~\citep{dwivedi2021graph,kreuzer2021rethinking} and iterative encoding of local sub-structures as auxiliary information~\citep{mialon2021graphit}. A recent study~\citep{rampavsek2022recipe} presents a generalized recipe for developing GTs. Despite the promising results of GTs in general transfer learning tasks, no current studies have successfully applied them to the unique challenge of few-shot learning scenarios with limited labeled data.

\subsection{Learning with Prompts on Graphs.} 
Prompting, as highlighted in~\citep{liu2021pre,lester2021power}, is a recent development in Natural Language Processing (NLP) that adapts large language models to various downstream tasks by adding task descriptions to input texts. This technique has inspired several studies~\citep{sun2022gppt,fang2022prompt,liu2023graphprompt} to apply such prompting methods to pretrained message-passing GNNs on symbolic graph data. For instance, GPF~\citep{fang2022prompt} introduces learnable perturbations as prompts for graph-level tasks. GPPT~\citep{sun2022gppt} and Graph Prompt~\citep{liu2023graphprompt} propose a uniform template for pretext tasks and targeted downstream tasks to facilitate prompting. Contrasting with these works, our framework introduces adjustable virtual nodes in the embedding space of Graph Transformers. By incorporating a unique GPPE module, our framework can address scenarios with sparse labels in base classes. Our paper is the first to present a prompt-based method for GTs, specifically designed for few-shot node classification tasks.

\section{Conclusion}
In this paper, we propose a novel approach, dubbed as Virtual Node Tuning (VNT), to tackle the problem of general few-shot node classification (FSNC) on symbolic graphs. This method adjusts pretrained graph transformers (GTs) by incorporating virtual nodes in the embedding space for tailored node embeddings. We also design a Graph-based Pseudo Prompt Evolution (GPPE) module for efficient knowledge transfer in scenarios with sparse labels. Our comprehensive empirical studies showcase our method's effectiveness and its potential for prompt initialization and ensemble. Our research thus pioneers a novel approach for learning on graphs under limited supervision and fine-tuning GTs for a target domain.
\begin{acks}
This work is supported by the National Science Foundation (NSF) under grants IIS-2229461.
\end{acks}

\bibliographystyle{ACM-Reference-Format}
\balance
\bibliography{reference}


\begin{thebibliography}{52}


\ifx \showCODEN    \undefined \def \showCODEN     #1{\unskip}     \fi
\ifx \showDOI      \undefined \def \showDOI       #1{#1}\fi
\ifx \showISBNx    \undefined \def \showISBNx     #1{\unskip}     \fi
\ifx \showISBNxiii \undefined \def \showISBNxiii  #1{\unskip}     \fi
\ifx \showISSN     \undefined \def \showISSN      #1{\unskip}     \fi
\ifx \showLCCN     \undefined \def \showLCCN      #1{\unskip}     \fi
\ifx \shownote     \undefined \def \shownote      #1{#1}          \fi
\ifx \showarticletitle \undefined \def \showarticletitle #1{#1}   \fi
\ifx \showURL      \undefined \def \showURL       {\relax}        \fi
\providecommand\bibfield[2]{#2}
\providecommand\bibinfo[2]{#2}
\providecommand\natexlab[1]{#1}
\providecommand\showeprint[2][]{arXiv:#2}

\bibitem[Bojchevski and G{\"u}nnemann(2018)]%
        {bojchevski2018deep}
\bibfield{author}{\bibinfo{person}{Aleksandar Bojchevski} {and}
  \bibinfo{person}{Stephan G{\"u}nnemann}.} \bibinfo{year}{2018}\natexlab{}.
\newblock \showarticletitle{Deep Gaussian Embedding of Graphs: Unsupervised
  Inductive Learning via Ranking}. In \bibinfo{booktitle}{\emph{ICLR}}.
\newblock


\bibitem[Chen et~al\mbox{.}(2022)]%
        {chen2022structure}
\bibfield{author}{\bibinfo{person}{Dexiong Chen}, \bibinfo{person}{Leslie
  O’Bray}, {and} \bibinfo{person}{Karsten Borgwardt}.}
  \bibinfo{year}{2022}\natexlab{}.
\newblock \showarticletitle{Structure-aware transformer for graph
  representation learning}. In \bibinfo{booktitle}{\emph{International
  Conference on Machine Learning}}. PMLR, \bibinfo{pages}{3469--3489}.
\newblock


\bibitem[Devlin et~al\mbox{.}(2018)]%
        {devlin2018bert}
\bibfield{author}{\bibinfo{person}{Jacob Devlin}, \bibinfo{person}{Ming-Wei
  Chang}, \bibinfo{person}{Kenton Lee}, {and} \bibinfo{person}{Kristina
  Toutanova}.} \bibinfo{year}{2018}\natexlab{}.
\newblock \showarticletitle{Bert: Pre-training of deep bidirectional
  transformers for language understanding}.
\newblock \bibinfo{journal}{\emph{arXiv preprint arXiv:1810.04805}}
  (\bibinfo{year}{2018}).
\newblock


\bibitem[Ding et~al\mbox{.}(2020)]%
        {ding2020graph}
\bibfield{author}{\bibinfo{person}{Kaize Ding}, \bibinfo{person}{Jianling
  Wang}, \bibinfo{person}{Jundong Li}, \bibinfo{person}{Kai Shu},
  \bibinfo{person}{Chenghao Liu}, {and} \bibinfo{person}{Huan Liu}.}
  \bibinfo{year}{2020}\natexlab{}.
\newblock \showarticletitle{Graph prototypical networks for few-shot learning
  on attributed networks}. In \bibinfo{booktitle}{\emph{CIKM}}.
\newblock


\bibitem[Ding et~al\mbox{.}(2022)]%
        {ding2022data}
\bibfield{author}{\bibinfo{person}{Kaize Ding}, \bibinfo{person}{Zhe Xu},
  \bibinfo{person}{Hanghang Tong}, {and} \bibinfo{person}{Huan Liu}.}
  \bibinfo{year}{2022}\natexlab{}.
\newblock \showarticletitle{Data augmentation for deep graph learning: A
  survey}.
\newblock \bibinfo{journal}{\emph{arXiv preprint arXiv:2202.08235}}
  (\bibinfo{year}{2022}).
\newblock


\bibitem[Dosovitskiy et~al\mbox{.}(2020)]%
        {dosovitskiy2020image}
\bibfield{author}{\bibinfo{person}{Alexey Dosovitskiy}, \bibinfo{person}{Lucas
  Beyer}, \bibinfo{person}{Alexander Kolesnikov}, \bibinfo{person}{Dirk
  Weissenborn}, \bibinfo{person}{Xiaohua Zhai}, \bibinfo{person}{Thomas
  Unterthiner}, \bibinfo{person}{Mostafa Dehghani}, \bibinfo{person}{Matthias
  Minderer}, \bibinfo{person}{Georg Heigold}, \bibinfo{person}{Sylvain Gelly},
  {et~al\mbox{.}}} \bibinfo{year}{2020}\natexlab{}.
\newblock \showarticletitle{An image is worth 16x16 words: Transformers for
  image recognition at scale}.
\newblock \bibinfo{journal}{\emph{arXiv preprint arXiv:2010.11929}}
  (\bibinfo{year}{2020}).
\newblock


\bibitem[Dwivedi et~al\mbox{.}(2021)]%
        {dwivedi2021graph}
\bibfield{author}{\bibinfo{person}{Vijay~Prakash Dwivedi},
  \bibinfo{person}{Anh~Tuan Luu}, \bibinfo{person}{Thomas Laurent},
  \bibinfo{person}{Yoshua Bengio}, {and} \bibinfo{person}{Xavier Bresson}.}
  \bibinfo{year}{2021}\natexlab{}.
\newblock \showarticletitle{Graph neural networks with learnable structural and
  positional representations}.
\newblock \bibinfo{journal}{\emph{arXiv preprint arXiv:2110.07875}}
  (\bibinfo{year}{2021}).
\newblock


\bibitem[Fang et~al\mbox{.}(2022)]%
        {fang2022prompt}
\bibfield{author}{\bibinfo{person}{Taoran Fang}, \bibinfo{person}{Yunchao
  Zhang}, \bibinfo{person}{Yang Yang}, {and} \bibinfo{person}{Chunping Wang}.}
  \bibinfo{year}{2022}\natexlab{}.
\newblock \showarticletitle{Prompt Tuning for Graph Neural Networks}.
\newblock \bibinfo{journal}{\emph{arXiv preprint arXiv:2209.15240}}
  (\bibinfo{year}{2022}).
\newblock


\bibitem[Fey and Lenssen(2019)]%
        {Fey/Lenssen/2019}
\bibfield{author}{\bibinfo{person}{Matthias Fey} {and} \bibinfo{person}{Jan~E.
  Lenssen}.} \bibinfo{year}{2019}\natexlab{}.
\newblock \showarticletitle{Fast Graph Representation Learning with {PyTorch
  Geometric}}. In \bibinfo{booktitle}{\emph{ICLR Workshop on Representation
  Learning on Graphs and Manifolds}}.
\newblock


\bibitem[Finn et~al\mbox{.}(2017)]%
        {finn2017model}
\bibfield{author}{\bibinfo{person}{Chelsea Finn}, \bibinfo{person}{Pieter
  Abbeel}, {and} \bibinfo{person}{Sergey Levine}.}
  \bibinfo{year}{2017}\natexlab{}.
\newblock \showarticletitle{Model-agnostic meta-learning for fast adaptation of
  deep networks}. In \bibinfo{booktitle}{\emph{ICML}}.
\newblock


\bibitem[Hamilton et~al\mbox{.}(2017)]%
        {Hamilton:2017tp}
\bibfield{author}{\bibinfo{person}{William~L. Hamilton},
  \bibinfo{person}{Zhitao Ying}, {and} \bibinfo{person}{Jure Leskovec}.}
  \bibinfo{year}{2017}\natexlab{}.
\newblock \showarticletitle{{Inductive Representation Learning on Large
  Graphs}}. In \bibinfo{booktitle}{\emph{NeurIPS}}.
  \bibinfo{pages}{1024--1034}.
\newblock


\bibitem[Hassani and Khasahmadi(2020)]%
        {hassani2020contrastive}
\bibfield{author}{\bibinfo{person}{Kaveh Hassani} {and}
  \bibinfo{person}{Amir~Hosein Khasahmadi}.} \bibinfo{year}{2020}\natexlab{}.
\newblock \showarticletitle{Contrastive multi-view representation learning on
  graphs}. In \bibinfo{booktitle}{\emph{International Conference on Machine
  Learning}}. PMLR, \bibinfo{pages}{4116--4126}.
\newblock


\bibitem[Hu et~al\mbox{.}(2020b)]%
        {hu2020open}
\bibfield{author}{\bibinfo{person}{Weihua Hu}, \bibinfo{person}{Matthias Fey},
  \bibinfo{person}{Marinka Zitnik}, \bibinfo{person}{Yuxiao Dong},
  \bibinfo{person}{Hongyu Ren}, \bibinfo{person}{Bowen Liu},
  \bibinfo{person}{Michele Catasta}, {and} \bibinfo{person}{Jure Leskovec}.}
  \bibinfo{year}{2020}\natexlab{b}.
\newblock \showarticletitle{Open Graph Benchmark: Datasets for Machine Learning
  on Graphs}. In \bibinfo{booktitle}{\emph{NeurIPS}}.
\newblock


\bibitem[Hu et~al\mbox{.}(2020a)]%
        {hu2020heterogeneous}
\bibfield{author}{\bibinfo{person}{Ziniu Hu}, \bibinfo{person}{Yuxiao Dong},
  \bibinfo{person}{Kuansan Wang}, {and} \bibinfo{person}{Yizhou Sun}.}
  \bibinfo{year}{2020}\natexlab{a}.
\newblock \showarticletitle{Heterogeneous graph transformer}. In
  \bibinfo{booktitle}{\emph{Proceedings of The Web Conference 2020}}.
  \bibinfo{pages}{2704--2710}.
\newblock


\bibitem[Huang and Zitnik(2020)]%
        {huang2020graph}
\bibfield{author}{\bibinfo{person}{Kexin Huang} {and} \bibinfo{person}{Marinka
  Zitnik}.} \bibinfo{year}{2020}\natexlab{}.
\newblock \showarticletitle{Graph meta learning via local subgraphs}. In
  \bibinfo{booktitle}{\emph{NeurIPS}}.
\newblock


\bibitem[Jia et~al\mbox{.}(2022)]%
        {jia2022visual}
\bibfield{author}{\bibinfo{person}{Menglin Jia}, \bibinfo{person}{Luming Tang},
  \bibinfo{person}{Bor-Chun Chen}, \bibinfo{person}{Claire Cardie},
  \bibinfo{person}{Serge Belongie}, \bibinfo{person}{Bharath Hariharan}, {and}
  \bibinfo{person}{Ser-Nam Lim}.} \bibinfo{year}{2022}\natexlab{}.
\newblock \showarticletitle{Visual prompt tuning}.
\newblock \bibinfo{journal}{\emph{arXiv preprint arXiv:2203.12119}}
  (\bibinfo{year}{2022}).
\newblock


\bibitem[Jiao et~al\mbox{.}(2020)]%
        {jiao2020sub}
\bibfield{author}{\bibinfo{person}{Yizhu Jiao}, \bibinfo{person}{Yun Xiong},
  \bibinfo{person}{Jiawei Zhang}, \bibinfo{person}{Yao Zhang},
  \bibinfo{person}{Tianqi Zhang}, {and} \bibinfo{person}{Yangyong Zhu}.}
  \bibinfo{year}{2020}\natexlab{}.
\newblock \showarticletitle{Sub-graph contrast for scalable self-supervised
  graph representation learning}. In \bibinfo{booktitle}{\emph{2020 IEEE
  international conference on data mining (ICDM)}}. IEEE,
  \bibinfo{pages}{222--231}.
\newblock


\bibitem[Jin et~al\mbox{.}(2021)]%
        {jin2021multi}
\bibfield{author}{\bibinfo{person}{Ming Jin}, \bibinfo{person}{Yizhen Zheng},
  \bibinfo{person}{Yuan-Fang Li}, \bibinfo{person}{Chen Gong},
  \bibinfo{person}{Chuan Zhou}, {and} \bibinfo{person}{Shirui Pan}.}
  \bibinfo{year}{2021}\natexlab{}.
\newblock \showarticletitle{Multi-scale contrastive siamese networks for
  self-supervised graph representation learning}. In
  \bibinfo{booktitle}{\emph{International Joint Conference on Artificial
  Intelligence 2021}}. Association for the Advancement of Artificial
  Intelligence (AAAI), \bibinfo{pages}{1477--1483}.
\newblock


\bibitem[Kipf and Welling(2017)]%
        {Kipf:2017tc}
\bibfield{author}{\bibinfo{person}{Thomas~N. Kipf} {and} \bibinfo{person}{Max
  Welling}.} \bibinfo{year}{2017}\natexlab{}.
\newblock \showarticletitle{{Semi-Supervised Classification with Graph
  Convolutional Networks}}. In \bibinfo{booktitle}{\emph{ICLR}}.
\newblock


\bibitem[Kreuzer et~al\mbox{.}(2021)]%
        {kreuzer2021rethinking}
\bibfield{author}{\bibinfo{person}{Devin Kreuzer}, \bibinfo{person}{Dominique
  Beaini}, \bibinfo{person}{Will Hamilton}, \bibinfo{person}{Vincent
  L{\'e}tourneau}, {and} \bibinfo{person}{Prudencio Tossou}.}
  \bibinfo{year}{2021}\natexlab{}.
\newblock \showarticletitle{Rethinking graph transformers with spectral
  attention}.
\newblock \bibinfo{journal}{\emph{Advances in Neural Information Processing
  Systems}}  \bibinfo{volume}{34} (\bibinfo{year}{2021}),
  \bibinfo{pages}{21618--21629}.
\newblock


\bibitem[Lan et~al\mbox{.}(2020)]%
        {lan2020node}
\bibfield{author}{\bibinfo{person}{Lin Lan}, \bibinfo{person}{Pinghui Wang},
  \bibinfo{person}{Xuefeng Du}, \bibinfo{person}{Kaikai Song},
  \bibinfo{person}{Jing Tao}, {and} \bibinfo{person}{Xiaohong Guan}.}
  \bibinfo{year}{2020}\natexlab{}.
\newblock \showarticletitle{Node classification on graphs with few-shot novel
  labels via meta transformed network embedding}.
\newblock \bibinfo{journal}{\emph{Advances in Neural Information Processing
  Systems}}  \bibinfo{volume}{33} (\bibinfo{year}{2020}),
  \bibinfo{pages}{16520--16531}.
\newblock


\bibitem[Lester et~al\mbox{.}(2021)]%
        {lester2021power}
\bibfield{author}{\bibinfo{person}{Brian Lester}, \bibinfo{person}{Rami
  Al-Rfou}, {and} \bibinfo{person}{Noah Constant}.}
  \bibinfo{year}{2021}\natexlab{}.
\newblock \showarticletitle{The power of scale for parameter-efficient prompt
  tuning}.
\newblock \bibinfo{journal}{\emph{arXiv preprint arXiv:2104.08691}}
  (\bibinfo{year}{2021}).
\newblock


\bibitem[Liu et~al\mbox{.}(2021b)]%
        {liu2021pre}
\bibfield{author}{\bibinfo{person}{Pengfei Liu}, \bibinfo{person}{Weizhe Yuan},
  \bibinfo{person}{Jinlan Fu}, \bibinfo{person}{Zhengbao Jiang},
  \bibinfo{person}{Hiroaki Hayashi}, {and} \bibinfo{person}{Graham Neubig}.}
  \bibinfo{year}{2021}\natexlab{b}.
\newblock \showarticletitle{Pre-train, prompt, and predict: A systematic survey
  of prompting methods in natural language processing}.
\newblock \bibinfo{journal}{\emph{arXiv preprint arXiv:2107.13586}}
  (\bibinfo{year}{2021}).
\newblock


\bibitem[Liu et~al\mbox{.}(2021a)]%
        {liu2021relative}
\bibfield{author}{\bibinfo{person}{Zemin Liu}, \bibinfo{person}{Yuan Fang},
  \bibinfo{person}{Chenghao Liu}, {and} \bibinfo{person}{Steven~CH Hoi}.}
  \bibinfo{year}{2021}\natexlab{a}.
\newblock \showarticletitle{Relative and absolute location embedding for
  few-shot node classification on graph}. In \bibinfo{booktitle}{\emph{AAAI}}.
\newblock


\bibitem[Liu et~al\mbox{.}(2023)]%
        {liu2023graphprompt}
\bibfield{author}{\bibinfo{person}{Zemin Liu}, \bibinfo{person}{Xingtong Yu},
  \bibinfo{person}{Yuan Fang}, {and} \bibinfo{person}{Xinming Zhang}.}
  \bibinfo{year}{2023}\natexlab{}.
\newblock \showarticletitle{GraphPrompt: Unifying Pre-Training and Downstream
  Tasks for Graph Neural Networks}. In \bibinfo{booktitle}{\emph{Proceedings of
  the ACM Web Conference 2023}}. \bibinfo{pages}{417--428}.
\newblock


\bibitem[Mialon et~al\mbox{.}(2021)]%
        {mialon2021graphit}
\bibfield{author}{\bibinfo{person}{Gr{\'e}goire Mialon},
  \bibinfo{person}{Dexiong Chen}, \bibinfo{person}{Margot Selosse}, {and}
  \bibinfo{person}{Julien Mairal}.} \bibinfo{year}{2021}\natexlab{}.
\newblock \showarticletitle{Graphit: Encoding graph structure in transformers}.
\newblock \bibinfo{journal}{\emph{arXiv preprint arXiv:2106.05667}}
  (\bibinfo{year}{2021}).
\newblock


\bibitem[Mo et~al\mbox{.}(2022)]%
        {mo2022simple}
\bibfield{author}{\bibinfo{person}{Yujie Mo}, \bibinfo{person}{Liang Peng},
  \bibinfo{person}{Jie Xu}, \bibinfo{person}{Xiaoshuang Shi}, {and}
  \bibinfo{person}{Xiaofeng Zhu}.} \bibinfo{year}{2022}\natexlab{}.
\newblock \showarticletitle{Simple unsupervised graph representation learning}.
  AAAI.
\newblock


\bibitem[Pan et~al\mbox{.}(2021)]%
        {pan2021neural}
\bibfield{author}{\bibinfo{person}{Liming Pan}, \bibinfo{person}{Cheng Shi},
  {and} \bibinfo{person}{Ivan Dokmani{\'c}}.} \bibinfo{year}{2021}\natexlab{}.
\newblock \showarticletitle{Neural Link Prediction with Walk Pooling}. In
  \bibinfo{booktitle}{\emph{International Conference on Learning
  Representations}}.
\newblock


\bibitem[Ramp{\'a}{\v{s}}ek et~al\mbox{.}(2022)]%
        {rampavsek2022recipe}
\bibfield{author}{\bibinfo{person}{Ladislav Ramp{\'a}{\v{s}}ek},
  \bibinfo{person}{Michael Galkin}, \bibinfo{person}{Vijay~Prakash Dwivedi},
  \bibinfo{person}{Anh~Tuan Luu}, \bibinfo{person}{Guy Wolf}, {and}
  \bibinfo{person}{Dominique Beaini}.} \bibinfo{year}{2022}\natexlab{}.
\newblock \showarticletitle{Recipe for a general, powerful, scalable graph
  transformer}.
\newblock \bibinfo{journal}{\emph{Advances in Neural Information Processing
  Systems}}  \bibinfo{volume}{35} (\bibinfo{year}{2022}),
  \bibinfo{pages}{14501--14515}.
\newblock


\bibitem[Rong et~al\mbox{.}(2020)]%
        {rong2020self}
\bibfield{author}{\bibinfo{person}{Yu Rong}, \bibinfo{person}{Yatao Bian},
  \bibinfo{person}{Tingyang Xu}, \bibinfo{person}{Weiyang Xie},
  \bibinfo{person}{Ying Wei}, \bibinfo{person}{Wenbing Huang}, {and}
  \bibinfo{person}{Junzhou Huang}.} \bibinfo{year}{2020}\natexlab{}.
\newblock \showarticletitle{Self-supervised graph transformer on large-scale
  molecular data}.
\newblock \bibinfo{journal}{\emph{Advances in Neural Information Processing
  Systems}}  \bibinfo{volume}{33} (\bibinfo{year}{2020}),
  \bibinfo{pages}{12559--12571}.
\newblock


\bibitem[Snell et~al\mbox{.}(2017)]%
        {snell2017prototypical}
\bibfield{author}{\bibinfo{person}{Jake Snell}, \bibinfo{person}{Kevin
  Swersky}, {and} \bibinfo{person}{Richard Zemel}.}
  \bibinfo{year}{2017}\natexlab{}.
\newblock \showarticletitle{Prototypical networks for few-shot learning}. In
  \bibinfo{booktitle}{\emph{NeurIPS}}.
\newblock


\bibitem[Sun et~al\mbox{.}(2022)]%
        {sun2022gppt}
\bibfield{author}{\bibinfo{person}{Mingchen Sun}, \bibinfo{person}{Kaixiong
  Zhou}, \bibinfo{person}{Xin He}, \bibinfo{person}{Ying Wang}, {and}
  \bibinfo{person}{Xin Wang}.} \bibinfo{year}{2022}\natexlab{}.
\newblock \showarticletitle{Gppt: Graph pre-training and prompt tuning to
  generalize graph neural networks}. In \bibinfo{booktitle}{\emph{Proceedings
  of the 28th ACM SIGKDD Conference on Knowledge Discovery and Data Mining}}.
  \bibinfo{pages}{1717--1727}.
\newblock


\bibitem[Tan et~al\mbox{.}(2022a)]%
        {tan2022graph}
\bibfield{author}{\bibinfo{person}{Zhen Tan}, \bibinfo{person}{Kaize Ding},
  \bibinfo{person}{Ruocheng Guo}, {and} \bibinfo{person}{Huan Liu}.}
  \bibinfo{year}{2022}\natexlab{a}.
\newblock \showarticletitle{Graph few-shot class-incremental learning}. In
  \bibinfo{booktitle}{\emph{WSDM}}.
\newblock


\bibitem[Tan et~al\mbox{.}(2022b)]%
        {tan2022simple}
\bibfield{author}{\bibinfo{person}{Zhen Tan}, \bibinfo{person}{Kaize Ding},
  \bibinfo{person}{Ruocheng Guo}, {and} \bibinfo{person}{Huan Liu}.}
  \bibinfo{year}{2022}\natexlab{b}.
\newblock \showarticletitle{A Simple Yet Effective Pretraining Strategy for
  Graph Few-shot Learning}.
\newblock \bibinfo{journal}{\emph{arXiv preprint arXiv:2203.15936}}
  (\bibinfo{year}{2022}).
\newblock


\bibitem[Tan et~al\mbox{.}(2022c)]%
        {tan2022transductive}
\bibfield{author}{\bibinfo{person}{Zhen Tan}, \bibinfo{person}{Song Wang},
  \bibinfo{person}{Kaize Ding}, \bibinfo{person}{Jundong Li}, {and}
  \bibinfo{person}{Huan Liu}.} \bibinfo{year}{2022}\natexlab{c}.
\newblock \showarticletitle{Transductive Linear Probing: A Novel Framework for
  Few-Shot Node Classification}.
\newblock \bibinfo{journal}{\emph{arXiv preprint arXiv:2212.05606}}
  (\bibinfo{year}{2022}).
\newblock


\bibitem[Thakoor et~al\mbox{.}(2021)]%
        {thakoor2021bootstrapped}
\bibfield{author}{\bibinfo{person}{Shantanu Thakoor}, \bibinfo{person}{Corentin
  Tallec}, \bibinfo{person}{Mohammad~Gheshlaghi Azar}, \bibinfo{person}{Remi
  Munos}, \bibinfo{person}{Petar Veli{\v{c}}kovi{\'c}}, {and}
  \bibinfo{person}{Michal Valko}.} \bibinfo{year}{2021}\natexlab{}.
\newblock \showarticletitle{Bootstrapped Representation Learning on Graphs}. In
  \bibinfo{booktitle}{\emph{ICLR Workshop on Geometrical and Topological
  Representation Learning}}.
\newblock


\bibitem[Vaswani et~al\mbox{.}(2017)]%
        {vaswani2017attention}
\bibfield{author}{\bibinfo{person}{Ashish Vaswani}, \bibinfo{person}{Noam
  Shazeer}, \bibinfo{person}{Niki Parmar}, \bibinfo{person}{Jakob Uszkoreit},
  \bibinfo{person}{Llion Jones}, \bibinfo{person}{Aidan~N Gomez},
  \bibinfo{person}{{\L}ukasz Kaiser}, {and} \bibinfo{person}{Illia
  Polosukhin}.} \bibinfo{year}{2017}\natexlab{}.
\newblock \showarticletitle{Attention is all you need}.
\newblock \bibinfo{journal}{\emph{Advances in neural information processing
  systems}}  \bibinfo{volume}{30} (\bibinfo{year}{2017}).
\newblock


\bibitem[Veli{\v c}kovi{\'c} et~al\mbox{.}(2018)]%
        {Velickovic:2018we}
\bibfield{author}{\bibinfo{person}{Petar Veli{\v c}kovi{\'c}},
  \bibinfo{person}{Guillem Cucurull}, \bibinfo{person}{Arantxa Casanova},
  \bibinfo{person}{Adriana Romero}, \bibinfo{person}{Pietro Li{\`o}}, {and}
  \bibinfo{person}{Yoshua Bengio}.} \bibinfo{year}{2018}\natexlab{}.
\newblock \showarticletitle{{Graph Attention Networks}}. In
  \bibinfo{booktitle}{\emph{ICLR}}.
\newblock


\bibitem[Wang et~al\mbox{.}(2020b)]%
        {wang2020microsoft}
\bibfield{author}{\bibinfo{person}{Kuansan Wang}, \bibinfo{person}{Zhihong
  Shen}, \bibinfo{person}{Chiyuan Huang}, \bibinfo{person}{Chieh-Han Wu},
  \bibinfo{person}{Yuxiao Dong}, {and} \bibinfo{person}{Anshul Kanakia}.}
  \bibinfo{year}{2020}\natexlab{b}.
\newblock \showarticletitle{Microsoft academic graph: When experts are not
  enough}.
\newblock \bibinfo{journal}{\emph{Quantitative Science Studies}}
  (\bibinfo{year}{2020}).
\newblock


\bibitem[Wang et~al\mbox{.}(2019)]%
        {wang2019dgl}
\bibfield{author}{\bibinfo{person}{Minjie Wang}, \bibinfo{person}{Da Zheng},
  \bibinfo{person}{Zihao Ye}, \bibinfo{person}{Quan Gan},
  \bibinfo{person}{Mufei Li}, \bibinfo{person}{Xiang Song},
  \bibinfo{person}{Jinjing Zhou}, \bibinfo{person}{Chao Ma},
  \bibinfo{person}{Lingfan Yu}, \bibinfo{person}{Yu Gai},
  \bibinfo{person}{Tianjun Xiao}, \bibinfo{person}{Tong He},
  \bibinfo{person}{George Karypis}, \bibinfo{person}{Jinyang Li}, {and}
  \bibinfo{person}{Zheng Zhang}.} \bibinfo{year}{2019}\natexlab{}.
\newblock \showarticletitle{Deep Graph Library: A Graph-Centric,
  Highly-Performant Package for Graph Neural Networks}.
\newblock \bibinfo{journal}{\emph{arXiv preprint arXiv:1909.01315}}
  (\bibinfo{year}{2019}).
\newblock


\bibitem[Wang et~al\mbox{.}(2020a)]%
        {wang21AMM}
\bibfield{author}{\bibinfo{person}{Ning Wang}, \bibinfo{person}{Minnan Luo},
  \bibinfo{person}{Kaize Ding}, \bibinfo{person}{Lingling Zhang},
  \bibinfo{person}{Jundong Li}, {and} \bibinfo{person}{Qinghua Zheng}.}
  \bibinfo{year}{2020}\natexlab{a}.
\newblock \showarticletitle{Graph Few-Shot Learning with Attribute Matching}.
  In \bibinfo{booktitle}{\emph{Proceedings of the 29th ACM International
  Conference on Information and Knowledge Management}}.
\newblock


\bibitem[Wang et~al\mbox{.}(2022)]%
        {wang2022task}
\bibfield{author}{\bibinfo{person}{Song Wang}, \bibinfo{person}{Kaize Ding},
  \bibinfo{person}{Chuxu Zhang}, \bibinfo{person}{Chen Chen}, {and}
  \bibinfo{person}{Jundong Li}.} \bibinfo{year}{2022}\natexlab{}.
\newblock \showarticletitle{Task-Adaptive Few-shot Node Classification}.
\newblock \bibinfo{journal}{\emph{arXiv preprint arXiv:2206.11972}}
  (\bibinfo{year}{2022}).
\newblock


\bibitem[Wen et~al\mbox{.}(2021)]%
        {wen2021meta}
\bibfield{author}{\bibinfo{person}{Zhihao Wen}, \bibinfo{person}{Yuan Fang},
  {and} \bibinfo{person}{Zemin Liu}.} \bibinfo{year}{2021}\natexlab{}.
\newblock \showarticletitle{Meta-inductive node classification across graphs}.
  In \bibinfo{booktitle}{\emph{Proceedings of the 44th International ACM SIGIR
  Conference on Research and Development in Information Retrieval}}.
\newblock


\bibitem[Wolf et~al\mbox{.}(2019)]%
        {wolf2019huggingface}
\bibfield{author}{\bibinfo{person}{Thomas Wolf}, \bibinfo{person}{Lysandre
  Debut}, \bibinfo{person}{Victor Sanh}, \bibinfo{person}{Julien Chaumond},
  \bibinfo{person}{Clement Delangue}, \bibinfo{person}{Anthony Moi},
  \bibinfo{person}{Pierric Cistac}, \bibinfo{person}{Tim Rault},
  \bibinfo{person}{R{\'e}mi Louf}, \bibinfo{person}{Morgan Funtowicz},
  {et~al\mbox{.}}} \bibinfo{year}{2019}\natexlab{}.
\newblock \showarticletitle{Huggingface's transformers: State-of-the-art
  natural language processing}.
\newblock \bibinfo{journal}{\emph{arXiv preprint arXiv:1910.03771}}
  (\bibinfo{year}{2019}).
\newblock


\bibitem[Xu et~al\mbox{.}(2019)]%
        {xu2018powerful}
\bibfield{author}{\bibinfo{person}{Keyulu Xu}, \bibinfo{person}{Weihua Hu},
  \bibinfo{person}{Jure Leskovec}, {and} \bibinfo{person}{Stefanie Jegelka}.}
  \bibinfo{year}{2019}\natexlab{}.
\newblock \showarticletitle{How powerful are graph neural networks?}. In
  \bibinfo{booktitle}{\emph{Proceedings of the 2019 International Conference on
  Learning Representations}}.
\newblock


\bibitem[Yang et~al\mbox{.}(2016)]%
        {yang2016revisiting}
\bibfield{author}{\bibinfo{person}{Zhilin Yang}, \bibinfo{person}{William
  Cohen}, {and} \bibinfo{person}{Ruslan Salakhudinov}.}
  \bibinfo{year}{2016}\natexlab{}.
\newblock \showarticletitle{Revisiting semi-supervised learning with graph
  embeddings}. In \bibinfo{booktitle}{\emph{International conference on machine
  learning}}. PMLR, \bibinfo{pages}{40--48}.
\newblock


\bibitem[Yao et~al\mbox{.}(2020)]%
        {yao2020graph}
\bibfield{author}{\bibinfo{person}{Huaxiu Yao}, \bibinfo{person}{Chuxu Zhang},
  \bibinfo{person}{Ying Wei}, \bibinfo{person}{Meng Jiang},
  \bibinfo{person}{Suhang Wang}, \bibinfo{person}{Junzhou Huang},
  \bibinfo{person}{Nitesh Chawla}, {and} \bibinfo{person}{Zhenhui Li}.}
  \bibinfo{year}{2020}\natexlab{}.
\newblock \showarticletitle{Graph few-shot learning via knowledge transfer}. In
  \bibinfo{booktitle}{\emph{Proceedings of the 34th AAAI Conference on
  Artificial Intelligence}}.
\newblock


\bibitem[You et~al\mbox{.}(2020)]%
        {you2020graph}
\bibfield{author}{\bibinfo{person}{Yuning You}, \bibinfo{person}{Tianlong
  Chen}, \bibinfo{person}{Yongduo Sui}, \bibinfo{person}{Ting Chen},
  \bibinfo{person}{Zhangyang Wang}, {and} \bibinfo{person}{Yang Shen}.}
  \bibinfo{year}{2020}\natexlab{}.
\newblock \showarticletitle{Graph contrastive learning with augmentations}.
\newblock \bibinfo{journal}{\emph{NeurIPS}} (\bibinfo{year}{2020}).
\newblock


\bibitem[Zhang et~al\mbox{.}(2020)]%
        {zhang2020graph}
\bibfield{author}{\bibinfo{person}{Jiawei Zhang}, \bibinfo{person}{Haopeng
  Zhang}, \bibinfo{person}{Congying Xia}, {and} \bibinfo{person}{Li Sun}.}
  \bibinfo{year}{2020}\natexlab{}.
\newblock \showarticletitle{Graph-bert: Only attention is needed for learning
  graph representations}.
\newblock \bibinfo{journal}{\emph{arXiv preprint arXiv:2001.05140}}
  (\bibinfo{year}{2020}).
\newblock


\bibitem[Zhang et~al\mbox{.}(2018)]%
        {zhang2018few}
\bibfield{author}{\bibinfo{person}{Shengzhong Zhang}, \bibinfo{person}{Ziang
  Zhou}, \bibinfo{person}{Zengfeng Huang}, {and} \bibinfo{person}{Zhongyu
  Wei}.} \bibinfo{year}{2018}\natexlab{}.
\newblock \showarticletitle{Few-shot Classification on Graphs with Structural
  Regularized GCNs}. In \bibinfo{booktitle}{\emph{Proceedings of the 32nd AAAI
  Conference on Artificial Intelligence}}.
\newblock


\bibitem[Zhou et~al\mbox{.}(2019)]%
        {zhou2019meta}
\bibfield{author}{\bibinfo{person}{Fan Zhou}, \bibinfo{person}{Chengtai Cao},
  \bibinfo{person}{Kunpeng Zhang}, \bibinfo{person}{Goce Trajcevski},
  \bibinfo{person}{Ting Zhong}, {and} \bibinfo{person}{Ji Geng}.}
  \bibinfo{year}{2019}\natexlab{}.
\newblock \showarticletitle{Meta-gnn: On few-shot node classification in graph
  meta-learning}. In \bibinfo{booktitle}{\emph{CIKM}}.
\newblock


\bibitem[Zhu et~al\mbox{.}(2020)]%
        {zhu2020deep}
\bibfield{author}{\bibinfo{person}{Yanqiao Zhu}, \bibinfo{person}{Yichen Xu},
  \bibinfo{person}{Feng Yu}, \bibinfo{person}{Qiang Liu}, \bibinfo{person}{Shu
  Wu}, {and} \bibinfo{person}{Liang Wang}.} \bibinfo{year}{2020}\natexlab{}.
\newblock \showarticletitle{Deep graph contrastive representation learning}.
\newblock \bibinfo{journal}{\emph{arXiv preprint arXiv:2006.04131}}
  (\bibinfo{year}{2020}).
\newblock


\end{thebibliography}

\newpage

\appendix

\section{Implementation Detail}
\label{app:implement}

\subsection{General Settings}
All experiments are implemented using PyTorch. We run all experiments on a single
80GB Nvidia A100 GPU.  
\subsection{Implementation of the Simplified GT}
For generality, we try to keep the used GT encoder very simple and easy to transfer to other complicated architectures. Specifically, we use a 1-layer MLP to project the raw graph, including node attributes and structural positions into the embedding space. We use simple summation to merge those embeddings together and feed them into the following transformer layers. We use the transformer module released by huggingface~\citep{wolf2019huggingface}. We perform a grid search like Section~\ref{exp:scale} to get the width and depth of the GT.

For pretraining, we utilize two prevailing pretext tasks, \textit{node attribute reconstruction} and \textit{structure recovery}, to train the GT encoder in a self-supervised manner. Concretely, for \textit{node attribute reconstruction} pretext, given a node, we minimize the \textit{Mean Square Error} (MSE) between the original node attributes and the reconstructed version via a fully connected layer and the learned node embedding from the GT. For \textit{structure recovery} pretext, given any pair of nodes, we try to predict if there is a link between them and compare the result with the ground truth by an MSE loss. To accommodate a larger graph, similar to~\citet{jiao2020sub,mo2022simple}, we adopt the mini-batch strategy to sample a portion of nodes with their subgraphs (based on PPR) in each epoch for pretraining.

\section{Details of Benchmark Datasets}
\label{app:stat}
\begin{table}[htbp]
	
	\setlength\tabcolsep{7.5pt}
\small
		\centering
		\renewcommand{\arraystretch}{1.6}

		\caption{Statistics of benchmark node classification datasets. $\sC_{train}$ denotes the base classes for training, $\sC_{dev}$ and $\sC_{test}$ denote novel classes for validation and test respectively.}
    \scalebox{0.7}{		
  \begin{tabular}{cccccccc}
		\hline
        \textbf{Dataset}&\# Nodes & \# Edges & \# Features &$|\mathbb{C}|$& $|\mathbb{C}_{train}|$& $|\mathbb{C}_{dev}|$& $|\mathbb{C}_{test}|$\\
        \hline

    \texttt{CoraFull}&    19,793&63,421&8,710&70&40&15&15\\\hline
 \texttt{Ogbn-arxiv}&169,343&1,166,243&128&40&20&10&10\\\hline
\texttt{Cora}&2,708&5,278&1,433&7&3&2&2\\\hline
 \texttt{CiteSeer}&3,327&4,552&3,703&6&2&2&2\\\hline

		\end{tabular}}
		\label{tab:statistics}
	\end{table}

\label{app:datasets}
In this section, we provide detailed descriptions of the benchmark datasets used in our experiments. 
All the datasets are public and available on both PyTorch-Geometric~\citep{Fey/Lenssen/2019} and DGL~\citep{wang2019dgl}.
\begin{itemize}
        \item \textbf{CoraFull}~\citep{bojchevski2018deep} is a citation network that extends the prevalent small Cora network. Specifically, it is achieved from the entire citation network, where nodes are papers, and edges denote the citation relations. The classes of nodes are obtained based on the paper topic.
    \item \textbf{Ogbn-arxiv}~\citep{hu2020open} is a directed citation network that consists of CS papers from MAG~\citep{wang2020microsoft}. Here nodes represent CS arXiv papers, and edges denote the citation relations. The classes of nodes are assigned based on the 40 subject areas of CS papers in arXiv. 
    \item \textbf{Cora}~\citep{yang2016revisiting} is a citation network dataset where nodes mean paper and edges mean citation relationships. Each node has a predefined feature with 1433 dimensions. The dataset is designed for the node classification task. The task is to predict the category of a certain paper.
    \item \textbf{CiteSeer}~\citep{yang2016revisiting} is also a citation network dataset where nodes mean scientific publications and edges mean citation relationships. Each node has a predefined feature with 3703 dimensions. The dataset is designed for the node classification task. The task is to predict the category of a certain publication.
    
\end{itemize}

\section{A More Detailed Review for Few-shot Node Classification}
\label{app:fsnc_review}
The task of few-shot node classification (FSNC) aims to train models that can assign labels to unlabeled nodes in graphs, using only a few labeled nodes per class for training. Recently, episodic meta-learning~\citep{finn2017model} has become a popular paradigm for addressing label scarcity in FSNC tasks. This approach trains GNN encoders by emulating the test environment for few-shot learning. For example, Meta-GNN \citep{zhou2019meta} uses MAML \citep{finn2017model} to learn optimization directions with limited labels. GPN~\citep{ding2020graph} employs Prototypical Networks \citep{snell2017prototypical} to perform classification based on the distance between node features and prototypes. MetaTNE~\citep{lan2020node} and RALE \citep{liu2021relative} also use episodic meta-learning to improve the adaptability of learned GNN encoders and achieve similar results. Additionally, G-Meta \citep{huang2020graph}, GFL-KT \citep{yao2020graph}, and MI-GNN~\citep{wen2021meta} use meta-learning to transfer knowledge when other auxiliary graphs are available. TNT~\citep{wang2022task} further takes into account the variance among different meta-tasks.

\section{Experiment Results of Design Discussion}\label{app:design}
\subsection{Prompt Initilization and Ensemble}
\begin{table}[htbp]
\vspace{-0.2cm}
\caption{\label{tab:initial_ensemble}The accuracy scores of VNT on \texttt{Cora} and \texttt{Ogbn-arxiv} datasets. Init. indicates the prototype-based prompt initialization strategy described in Section~\ref{sec:init}. Method without Init. means the prompts are randomly initialized. The best results are \textbf{bold}. MV refers to majority voting.}
\centering
\scalebox{0.56}{
\begin{tabular}{@{}ccc|cc|cccc@{}}
\toprule
\multicolumn{3}{c|}{VNT}                                                      & \multicolumn{2}{c|}{\texttt{Cora}}       & \multicolumn{4}{c}{\texttt{Ogbn-arxiv}}                                    \\ \midrule
\multicolumn{1}{c|}{\textbf{Init.}}     & \multicolumn{2}{c|}{\textbf{Ensemble}} & 2-way 1-shot   & 2-way 5-shot   & 2-way 1-shot   & 2-way 5-shot   & 5-way 1-shot   & 5-way 5-shot   \\ \midrule
\multicolumn{1}{c|}{}                   &                      &                 & 84.50          & 90.50          & 82.00          & 87.27          & 50.40          & 74.91          \\ \midrule
\multicolumn{1}{c|}{$\checkmark$}                  &                      &                 & 85.25          & 91.30          & 83.05          & 88.34          & 51.06          & 75.86          \\ \midrule
\multicolumn{1}{c|}{\multirow{3}{*}{$\checkmark$}} & \multirow{3}{*}{$\checkmark$}   & \textbf{Avg.}   & 85.50          & 89.64          & 82.50          & 89.65          & 48.82          & 75.65          \\
\multicolumn{1}{c|}{}                   &                      & \textbf{Best}   & 86.24          & 92.38          & \textbf{85.45} & 90.63          & \textbf{53.68} & 78.82          \\
\multicolumn{1}{c|}{}                   &                      & \textbf{MV}     & \textbf{87.63} & \textbf{92.52} & 84.72          & \textbf{91.50} & 52.15          & \textbf{79.60} \\ \bottomrule
\end{tabular}}
\end{table}

\subsection{Effectiveness with regard to the Scale of GTs}
\begin{figure}[htbp]
\vspace{-0.2cm}
\centering
\includegraphics[width=2.4in]{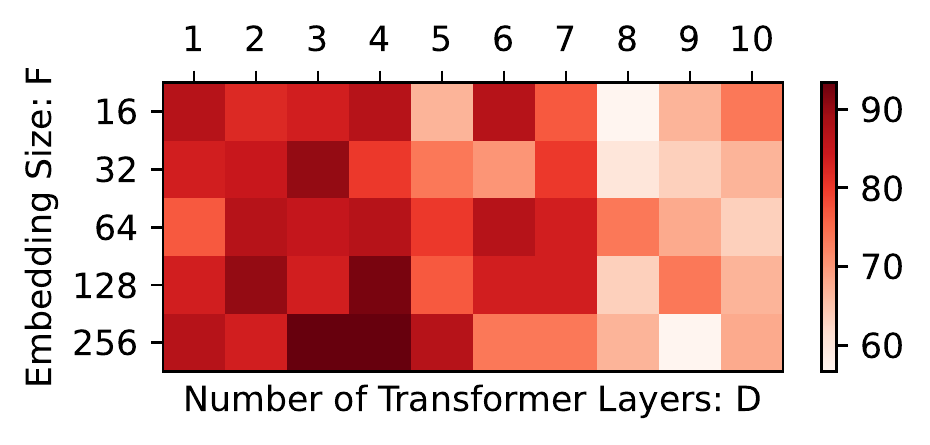}
\vspace{-0.2cm}
\caption{The $2$-way $1$-shot accuracy ($\%$) of the proposed VNT according to the scale of the GT encode on the \texttt{Cora} dataset.}
\label{fig:scale}
\vspace{-0.4cm}
\end{figure}



\section{Number of Virtual Nodes}\label{app:prompt_len}
\begin{figure}[htbp]
		\centering
		\scalebox{0.4}{
\includegraphics[width=0.98\textwidth]{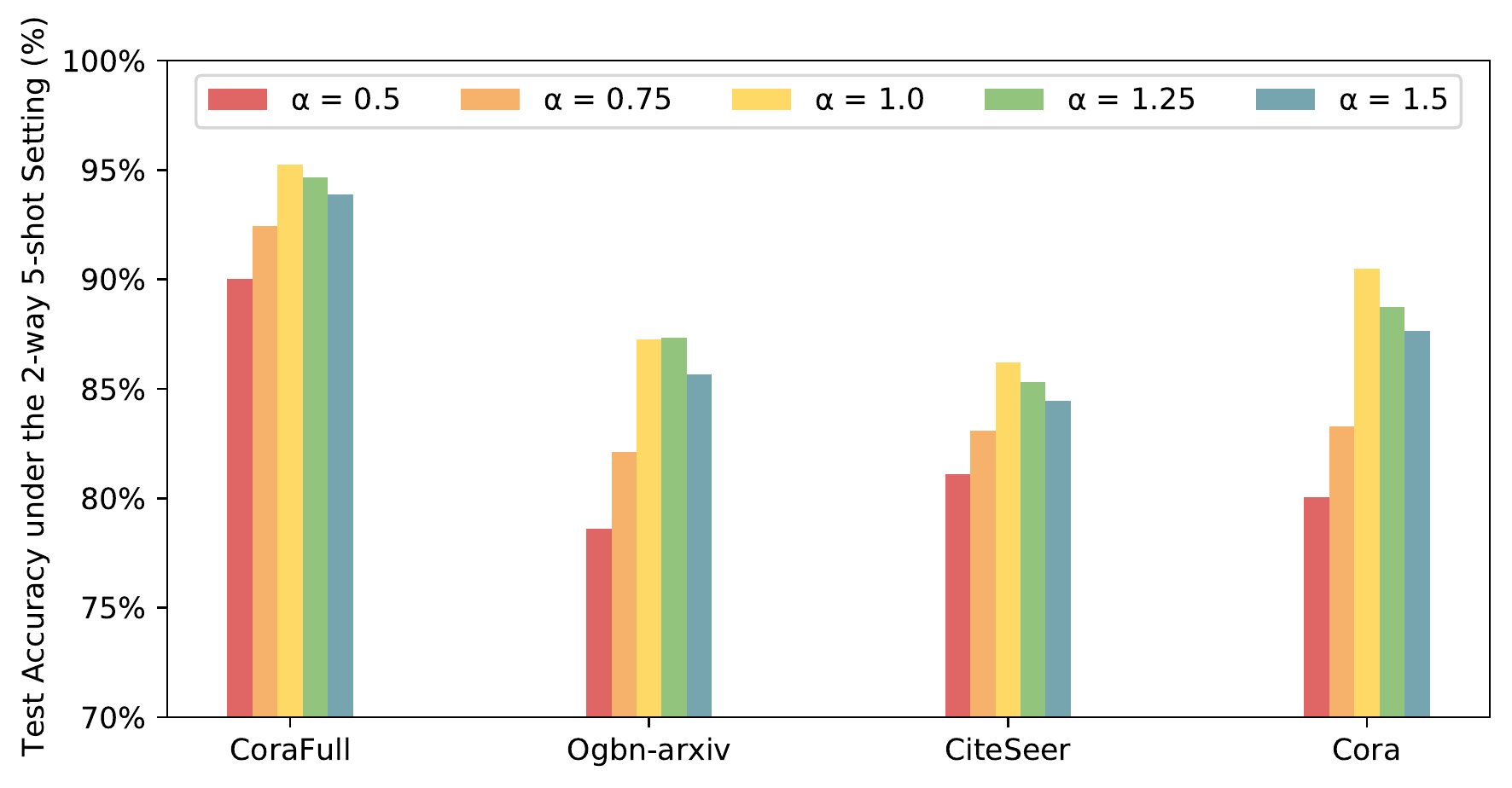}}
\caption{The test accuracy ($\%$) under different $\alpha$ values on four benchmark datasets.}
\label{fig:prompt_len}
	\end{figure}

The number of virtual nodes $P$ is an important hyper-parameter to tune. On the four benchmark datasets, we give the results under the $2$-way $5$-shot setting. Specifically, we use a parameter $\alpha$ to control the number of virtual nodes. For a $N$-way $K$-shot FSNC task, we define $\alpha = \frac{P}{N\cdot K}$. Larger $\alpha$ means more virtual nodes are introduced. From the results shown in Fig.~\ref{fig:prompt_len}, we find that when $\alpha = 1$, the proposed VNT can give the best performance. Therefore, we choose $P = \alpha \times (N\cdot K) = N\cdot K$ as the default number of virtual nodes per prompt.

\section{More Results on Node Clustering}
In table~\ref{tab:app-cluster}, we present the complete results on Node Clustering in terms of NMI and ARI.

\begin{table}[htbp]
		\setlength\tabcolsep{9.5pt}
	\small
		\centering
		\renewcommand{\arraystretch}{1.2}
		\caption{\label{tab:app-cluster}The overall NMI ($\uparrow$) and ARI ($\uparrow$) scores of baselines and ablated variants of the proposed framework on \texttt{CoraFull} and \texttt{CiteSeer} datasets. The best results among the variants and baselines are \textbf{bold} and \underline{underlined}, respectively.}
        \scalebox{1.}{
		\begin{tabular}{c||c|c||c|c}
			\hline
			\textbf{Dataset}&\multicolumn{2}{c||}{\texttt{CoraFull}}&\multicolumn{2}{c}{\texttt{CiteSeer}}
			\\
			\hline
						\textbf{Metrics}&\multicolumn{1}{c|}{\textbf{NMI}}&\multicolumn{1}{c||}{\textbf{ARI}}&\multicolumn{1}{c|}{\textbf{NMI}}&\multicolumn{1}{c}{\textbf{ARI}}\\

					\hline
     \multicolumn{5}{c}{Meta-learning} \\\hline
     MAML&$0.1622$&$0.0597$&$0.0754$&$0.0602$\\\hline		ProtoNet&$0.2669$&$0.1263$&$0.0915$&$0.0765$\\\hline
			AMM-GNN&$0.6247$&$0.5087$&$0.2090$&$0.1781$\\\hline
			G-Meta&$0.5003$&$0.3702$&$0.1913$&$0.1502$\\\hline
			Meta-GNN&$0.5534$&$0.4196$&$0.1317$&$0.1171$\\\hline
			GPN&$0.6001$&$0.4599$&$0.2119$&$0.2087$\\\hline
			TENT&$0.5760$&$0.4652$&$0.0930$&$0.0811$\\\hline

\multicolumn{5}{c}{GCL-based TLP} \\\hline
   
MVGRL&$0.6227$&$0.4788$&$0.2554$&$0.2232$\\\hline
GraphCL&${0.7023}$&${0.5628}$&$\underline{0.5579}$&$\underline{0.5890}$\\\hline
GRACE&$0.6781$&${0.5856}$&$0.2663$&$0.2778$\\\hline
BGRL&$0.5137$&$0.4382$&$0.2051$&$0.1875$\\\hline
MERIT&$0.7419$&$0.6590$&$0.3923$&$0.4014$\\\hline
SUGRL&$\underline{0.7680}$&$\underline{0.7049}$&$0.3952$&$0.4460$\\\hline

\multicolumn{5}{c}{Ablated variants of VNT} \\\hline

GT&$0.5225$&$0.3864$&$0.3452$&$0.3189$\\\hline
VNT&${0.7768}$&${0.6427}$&${0.5998}$&${0.6331}$\\\hline
VNT-GPPE&$\mathbf{0.7927}$&$\mathbf{0.7075}$&$\mathbf{0.6324}$&$\mathbf{0.6762}$\\\hline

	\hline
				
\end{tabular}}
		\label{tab:app_nmi_result}
	\end{table}

\label{app:clustering}


\end{document}